\documentclass[10pt,twocolumn,letterpaper]{article}

\usepackage[accsupp]{axessibility}
\usepackage[pagenumbers]{iccv} 

\def\ie{\emph{i.e.}\xspace}

\newcommand{\Fig}[1]{Fig.~\ref{fig:#1}}

\newcommand{\Subsec}[1]{Sec.~\ref{subsec:#1}}
\newcommand{\Eq}[1]{Eq.~(\ref{eq:#1})}
\newcommand{\Tbl}[1]{Tab.~\ref{tab:#1}}

\definecolor{iccvblue}{rgb}{0.21,0.49,0.74}
\usepackage[pagebackref,breaklinks,colorlinks,allcolors=iccvblue]{hyperref}
\usepackage{xfrac}
\usepackage{adjustbox}
\usepackage{multirow}
\usepackage{colortbl}
\usepackage{subcaption}
\usepackage{amsmath}
\def\confName{ICCV}
\def\confYear{2025}

\title{Axis-level Symmetry Detection with Group-Equivariant Representation}

\author{Wongyun Yu \quad \quad Ahyun Seo \quad \quad Minsu Cho\vspace{0.15cm}\\
Pohang University of Science and Technology (POSTECH), South Korea\\
{\tt\small \{yuwongyun, ahyun.seo, mscho\}@postech.ac.kr}\\
{\small \url{https://wongyun-yu.github.io/axis_symdet/}}
}

\begin{document}

\maketitle
\vspace{-200mm}
\begin{abstract}
Symmetry is a fundamental concept that has been extensively studied, yet detecting it in complex scenes remains a significant challenge in computer vision. Recent heatmap-based approaches can localize potential regions of symmetry axes but often lack precision in identifying individual axes. In this work, we propose a novel framework for axis-level detection of the two most common symmetry types—reflection and rotation—by representing them as explicit geometric primitives, \ie, lines and points. Our method employs a dual-branch architecture that is equivariant to the dihedral group, with each branch specialized to exploit the structure of dihedral group-equivariant features for its respective symmetry type. For reflection symmetry, we introduce \textbf{orientational anchors}, aligned with group components, to enable orientation-specific detection, and a \textbf{reflectional matching} that measures similarity between patterns and their mirrored counterparts across candidate axes. For rotational symmetry, we propose a \textbf{rotational matching} that compares patterns at fixed angular intervals to identify rotational centers. Extensive experiments demonstrate that our method achieves state-of-the-art performance, outperforming existing approaches.
\end{abstract}    
\begin{figure}[t]
    \centering
    \begin{tabular}{c}
        \includegraphics[width=0.9\columnwidth]{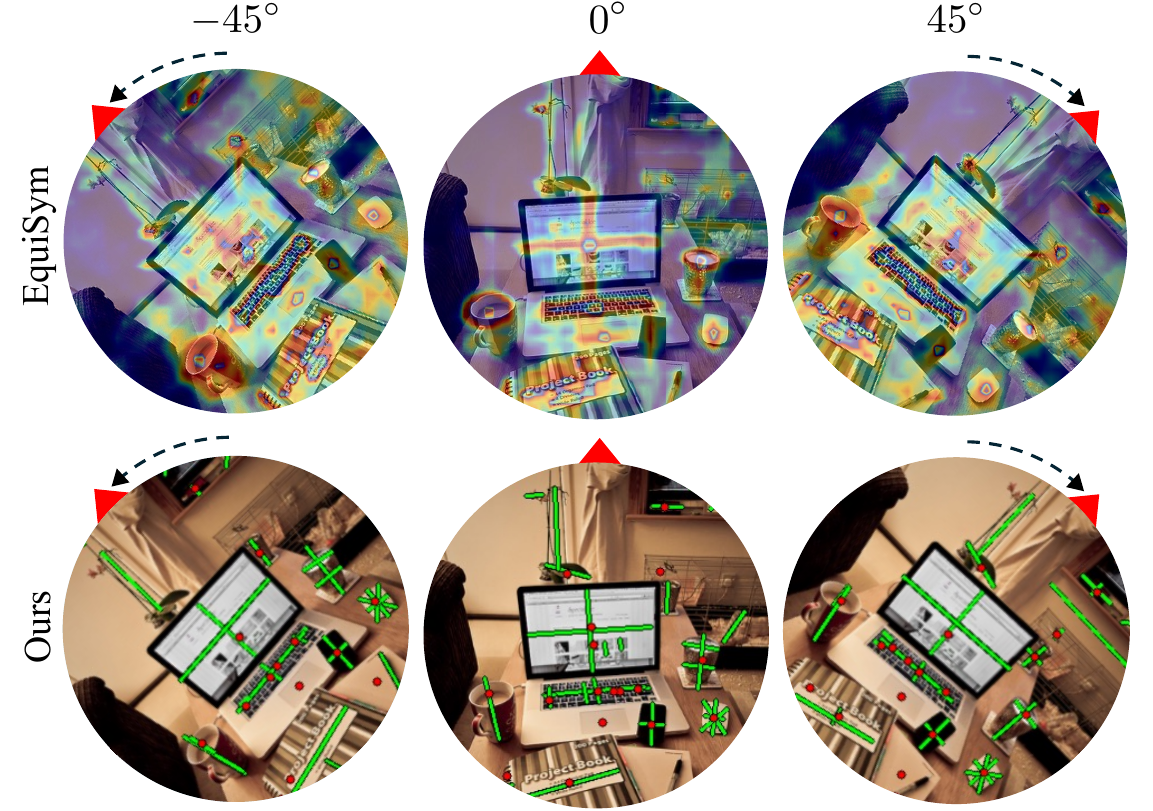} \\ 
    \end{tabular}
    \caption{
    \textbf{Comparison of the heatmap-based method~\cite{seo2022equisym} and our axis-level approach on rotated inputs.}
    The red triangle indicates the rotated orientation of input image. Our axis-level symmetry detection method captures reflection (green lines) and rotation (red points) axes as precise geometric entities and demonstrates superior robustness to rotation compared to the heatmap-based method.}
    \label{fig:teaser}
    \vspace{-6mm}
\end{figure}

\vspace{-5mm}
\section{Introduction}
Symmetry is a fundamental concept observed across natural and artificial environments~\cite{wertheimer1938laws, tyler1995empirical}, appearing at multiple scales and orientations~\cite{moller1998bilateral, giurfa1996symmetry}. While humans easily recognize symmetry~\cite{wagemans1995detection}, it remains a challenge for computer vision. This work focuses on detecting reflection and rotation symmetries in complex 2D scenes~\cite{liu2010computational}.  
Robust symmetry detection requires precise localization of symmetry axes, along with accurate determination of additional properties; reflection symmetry involves estimating axis length and orientation~\cite{zhang2021elsd}, while rotation symmetry requires classifying the correct fold. These challenges are amplified by real-world complexities such as occlusion and distortions.

Symmetry detection has evolved from classical to deep learning approaches. Traditional methods use descriptor matching for reflection symmetry~\cite{lowe2004distinctive, cho2009bilateral, loy2006detecting, kiryati1998detecting}, and frequency analysis~\cite{keller2006signal, lee2008rotation} or gradient flow~\cite{prasad2005detecting} for rotation symmetry. Neural networks advanced the field from early symmetry-aware models~\cite{fukushima2006symmetry, tsogkas2012learning}, to CNN-based prediction~\cite{funk2017beyond, teo2015detection}, and recent self-similarity and group-equivariant networks~\cite{seoshim2021pmcnet, seo2022equisym}.

Despite these advancements, recent neural network-based approaches face two key limitations.  First, most treat symmetry detection as a per-pixel heatmap prediction problem~\cite{funk2017beyond, seoshim2021pmcnet, seo2022equisym}, which makes it difficult to recover the precise geometric parameters of symmetry axes. Second, they either do not incorporate symmetric structure explicitly into feature representations ~\cite{seoshim2021pmcnet} or lack dedicated matching mechanisms ~\cite{seo2022equisym}, resulting in inconsistent outputs under image rotations or reflections.

To address these limitations, we propose an \emph{axis-level} symmetry detection that is equivariant to the dihedral group. We explicitly model reflection and rotation symmetries as geometric primitives—\ie, lines and points—and use features that are equivariant under the $\mathrm{D}_N$ (which includes planar rotations and reflections). Specialized modules are introduced for each symmetry type to ensure predictions transform predictably under input rotations and reflections.

For reflection symmetry, we propose \textbf{orientational anchors} aligned with the discrete orientations of dihedral group $\mathrm{D}_N$, enabling orientation-specific reflection axis detection. We also introduce a \textbf{reflectional matching} module that compares patterns with their mirrored counterparts across candidate axes. These modules are designed to be equivariant to $\mathrm{D}_N$, while being invariant under input reflections, consistent with the definition of reflection symmetry.

For rotation symmetry, we introduce a \textbf{rotational matching} module that compares features with rotated versions of themselves at fixed angles. This module is constructed to be fully invariant to dihedral group, ensuring consistent detection of rotational symmetry centers regardless of image rotation or reflection. \Fig{teaser} provides an overview, illustrating that our method produces consistent reflection and rotation axis detections under input transformations, unlike heatmap-based methods. Experiments on real-world datasets demonstrate that our method consistently outperforms existing pixel-level approaches in both reflection and rotation symmetry detection. 

The contributions of this paper include:
\begin{itemize}
    \item We propose a novel \textit{axis-level} symmetry detection network for reflection and rotation symmetry, leveraging representations equivariant to the dihedral group $\mathrm{D}_N$.
    \item We introduce an \textit{orientational anchor expansion} mechanism that enables orientation-specific detection by incorporating the group’s rotation dimension.
    \item We develop an \textit{equivariant reflectional matching} module and an \textit{invariant rotational matching} module for symmetry-consistent feature comparison.
    \item We validate our approach on real-world datasets and demonstrate superior performance compared to prior methods.
\end{itemize}

\section{Related work}
\paragraph{Symmetry detection.} 
Early reflection symmetry detection used keypoint matching~\cite{loy2006detecting, cho2009bilateral} with SIFT descriptors~\cite{lowe2004distinctive}, while contour~\cite{shen2001robust, wang2014unified} and gradient-based~\cite{sun1995symmetry, gnutti2021combining} methods extracted symmetry structures. Randomized approaches~\cite{cicconet2017finding} aligned patterns via cross-correlation, while Hough voting~\cite{cornelius2006detecting}, local affine frames~\cite{cornelius2007efficient}, and RANSAC~\cite{sinha2012detecting} refined axis extraction for planar surfaces.  
For rotation symmetry, early methods identified periodic signals in spatial~\cite{liu2004computational} and frequency domains~\cite{keller2006signal, lee2008rotation}, leveraging spectral density and angular correlation. SIFT-based techniques~\cite{lowe2004distinctive, loy2006detecting} normalized orientation for rotation detection, while GVF~\cite{prasad2005detecting} and polar domain representations~\cite{akbar2023detecting} improved boundary detection. Rectification methods~\cite{lee2009skewed} further addressed affine distortions.

Deep learning advanced symmetry detection from early feature extraction~\cite{fukushima2006symmetry, tsogkas2012learning} to CNN-based heatmaps~\cite{funk2017beyond, teo2015detection, seoshim2021pmcnet}, but focused on dense predictions. Recent optimization-based~\cite{je2024robust} and neural 3D symmetry reconstruction~\cite{li2024symmetry, Zhang_2023_ICCV} methods predict axis-level reflection symmetry but are limited to isolated objects without background context. \cite{seo2025leveraging} leverages 3D information for 2D axis-level symmetry detection but is limited to rotation symmetry. Feature matching remains underexplored—PMCNet~\cite{seoshim2021pmcnet} introduced polar matching but lacked explicit symmetry integration, while group-equivariant~\cite{gens2014deep, seo2022equisym} and invariant~\cite{dieleman2015rotation} architectures improved robustness but focused on appearance features.  
To address these limitations, we propose an architecture that detects both reflection and rotation symmetries at the scene level by modeling symmetry axes as geometric entities and integrating equivariant matching for symmetry-aware feature comparisons.

\paragraph{Equivariant neural networks.} 
Convolutional neural networks (CNNs) provide translation equivariance but lack rotation and reflection equivariance, limiting their effectiveness in symmetry-aware tasks. Group-equivariant CNNs~\cite{cohen2016group, cohen2016steerable, gens2014deep} introduced group convolutions to address this, with advancements in circular harmonics~\cite{worrall2017harmonic}, vector fields~\cite{marcos2017rotation}, and hexagonal lattices~\cite{hoogeboom2018hexaconv}. Later work extended equivariance to 3D data~\cite{weiler20183d}, intertwiner spaces~\cite{cohen2018intertwiners}, and homogeneous spaces~\cite{cohen2018general, cohen2019gauge, e2cnn}.
Equivariant models have been applied to aerial object detection~\cite{han2021redet} and symmetry detection~\cite{seo2022equisym}, with recent works improving keypoint descriptions~\cite{Bokman_2024_CVPR} and enforcing group-equivariant constraints for denoising~\cite{Terris_2024_CVPR}. Beyond 2D, 3D-equivariant architectures address pose estimation~\cite{lee20243d} and leverage spherical harmonics for 3D rotation-equivariant encoding~\cite{xu2025se}. Our approach builds upon dihedral-group equivariant networks, with specialized matching for enhanced symmetry detection.

\section{Background}
\paragraph{Group.}  
A group is a mathematical structure with a set and an operation satisfying closure, associativity, identity, and invertibility~\cite{rotman2012introduction}. Groups describe symmetries: transformations like rotations and reflections that preserve an object's structure. Our work is built upon two common discrete groups in neural networks: the cyclic group and the dihedral group.
The cyclic group $\mathrm{C}_N$ represents discrete rotations $\{r^0, \dots, r^{N-1}\}$, with the group law $r^i r^j = r^{(i+j) \bmod N}$. The dihedral group $\mathrm{D}_N$, relevant to our work, includes both rotations and reflections:  
\begin{align}  
\mathrm{D}_N = \{r^0, r^1,\dots, r^{N-1}, b, br^1,\dots, br^{N-1}\},  
\end{align}  
where $r$ and $b$ are generators of the dihedral group corresponding to rotation and reflection, respectively, satisfying $b^2 = e$ and $r^n b = b r^{-n}$, with $e$ as the identity. We use the regular representation~\cite{cohen2016group} to encode group actions, where each group element is represented as a permutation matrix acting on a vector space. A detailed explanation of group-equivariant representations and their use in convolutional architectures is provided in Appendix~\ref{appendix A}.

\paragraph{Equivariance.}  
A function $f: \mathcal{X} \to \mathcal{Y}$ is equivariant if it commutes with a group action. Formally, for linear group representations $\sigma_1: G \to \text{GL}(\mathcal{X})$ and $\sigma_2: G \to \text{GL}(\mathcal{Y})$, equivariance is defined as:  
\begin{equation}
    f(\sigma_1(g) \cdot x) = \sigma_2(g) \cdot f(x), \quad \forall g \in G, \, x \in \mathcal{X}.
\end{equation}
In neural networks, equivariance ensures that transformations in the input induce predictable transformations in the output, preserving data symmetries. 
\captionsetup[figure]{skip=2pt}
\begin{figure*}[t!]
    \centering
    \includegraphics[width=\textwidth]{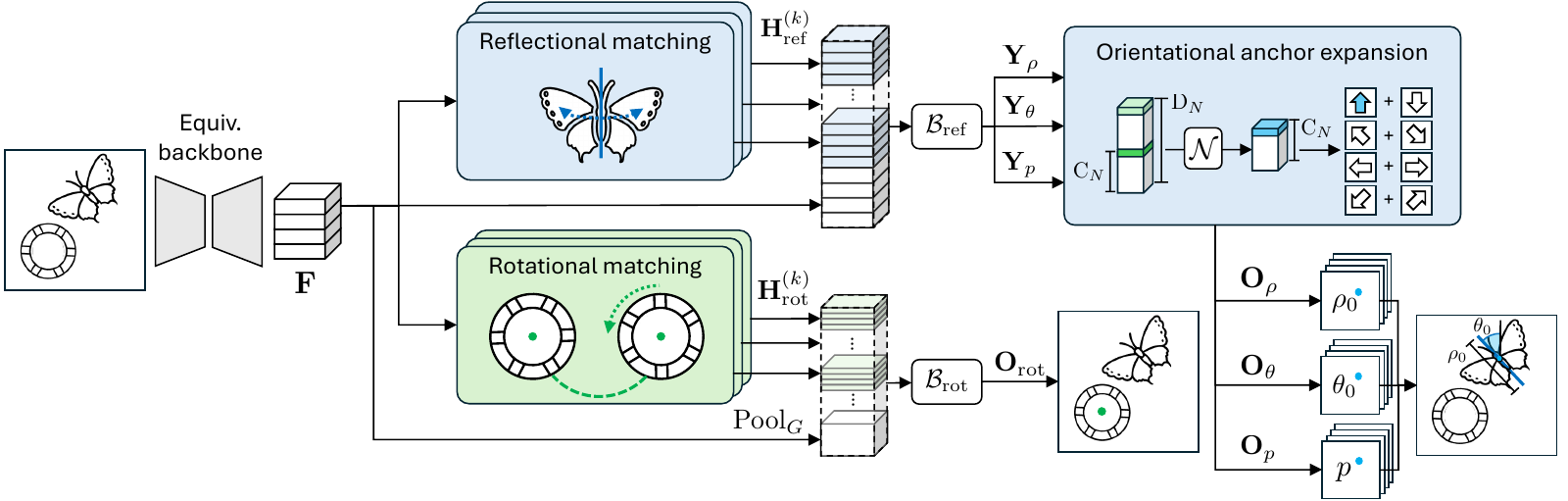}  
     \caption{\textbf{Overall architecture of our proposed instance-level symmetry detection network.} Given an input image, a $\mathrm{D}_N$-equivariant backbone extracts features $\mathbf{F} \in \mathbb{R}^{H \times W \times \mathcal{C} |\mathrm{D}_N|}$. The reflection branch (top) employs equivariant reflectional matching and orientational anchor expansion to predict reflection axes as parameterized line segments $(\alpha, x, y, p, \rho, \theta)$. The rotation branch (bottom) applies invariant rotational matching to detect rotation axes and classify their fold classes parameterized as $(x,y,p_s)$.}
    \label{fig:overall architecture 2col}
    \vspace{-4mm}
\end{figure*}

\section{Proposed method}
\label{sec:method}
In this section, we introduce a $\mathrm{D}_N$-equivariant network for axis-level symmetry detection, modeling reflection axes as line segments and rotation axes as points. The network uses a dihedral group-equivariant backbone~\cite{cohen2016group} to extract features, and then processes these features with two branches: one branch predicts the midpoint, orientation, and length of reflection axes, and the other predicts the location and fold class of rotation symmetry centers  (\Subsec{Instance-level Symmetry Detection}).
To handle multiple orientations, we introduce orientational anchor expansion, aligning feature channels with the discrete orientations of the dihedral group (\Subsec{Orientational Anchor Expansion}). We also present reflectional matching to capture symmetry across reflection axes (\Subsec{Reflectional Matching}) to compare features with their mirrored versions, and a rotational matching module (\Subsec{Rotational Matching}) to compare the same features at different rotation angles, while preserving dihedral group-equivariance. \Fig{overall architecture 2col} illustrates the overall pipeline of our approach. 

\subsection{Axis-level symmetry detection}
\label{subsec:Instance-level Symmetry Detection}  
Existing neural network-based methods detect symmetry in 2D scenes using pixel-level heatmaps~\cite{funk2017beyond, seoshim2021pmcnet, seo2022equisym}, methods detect symmetry by predicting dense pixel-wise heatmaps~\cite{funk2017beyond, seoshim2021pmcnet, seo2022equisym}, which makes it hard to recover exact axis parameters. Recent approaches represent reflection axes as lines~\cite{je2024robust, li2024symmetry} but are limited to isolated 2D or 3D objects without backgrounds.  
To address these limitations, we present an axis-level symmetry detection network that accurately represents reflection and rotation axes in complex real-world scenes with multiple instances.

\paragraph{Feature extraction.}
Given an input image $\mathbf{I}$, we employ a $\mathrm{D}_N$-equivariant backbone network~\cite{he2016deep, cohen2016group} to extract the base feature map $\mathbf{F} \in \mathbb{R}^{H \times W \times \mathcal{C} |\mathrm{D}_N|}$. Here, $H$ and $W$ denote the spatial dimensions, $|\mathrm{D}_N|=2N$ represents the number of dihedral group elements(combining $N$ rotations and their reflections), and $\mathcal{C}$ is the number of channels per group element. The extracted base feature $\mathbf{F}$ is then fed into the symmetry detection branches.

\paragraph{Reflection symmetry detection.}
For axis-level reflection symmetry detection, we model reflection axes using the center-angle-length representation~\cite{zhang2021elsd}.
Unlike the endpoint representation~\cite{zhou2019end, xue2020holistically, huang2018learning}, this approach inherently handles rotation and reflection equivariance via orientation parameterization. The reflection branch $\mathcal{B}_\mathrm{ref}$ processes the base feature map to predict the reflection axes components:
\begin{align}
    {\mathbf{Y}}_\text{ref} &= [{\mathbf{Y}}_p; {\mathbf{Y}}_\rho; {\mathbf{Y}}_\theta] = \mathcal{B}_\mathrm{ref}(\mathbf{F}) \in \mathbb{R}^{|\mathrm{D}_N| \times H \times W \times 3},\label{eq:reflection branch output}
\end{align}
where $\mathbf{Y}_p$ provides a probability for each spatial location being the midpoint of a reflection axis, and $\mathbf{Y}_\rho$ and $\mathbf{Y}_\theta$ are the regression outputs for the axis length and orientation at that location. 
To obtain reflection component map, we apply pooling across the group dimension, $\text{Pool}_G$ as follows:
\begin{align}
    \mathbf{O}_\text{ref} &= \text{Pool}_G({\mathbf{Y}}_\text{ref}) \in \mathbb{R}^{H \times W \times 3}.\label{pooled output}
\end{align}
At each position $(x, y)$, a reflection axis prediction is parameterized as $(x, y, p, \rho, \theta)$, where $p$ is the reflection axis midpoint probability, $\rho$ the length, and $\theta$ the orientation.
The start and end points of the predicted axis are given by:
\begin{align}
\begin{bmatrix}
    x_{\text{s}}\\
    y_{\text{s}}
\end{bmatrix}
&=
\begin{bmatrix}
    x \\
    y
\end{bmatrix}
+ \frac{\rho}{2} 
\begin{bmatrix}
    \cos(\theta) \\
    \sin(\theta)
\end{bmatrix}, \\
\begin{bmatrix}
    x_{\text{e}} \\
    y_{\text{e}}
\end{bmatrix}
&=
\begin{bmatrix}
    x \\
    y
\end{bmatrix}
- \frac{\rho}{2}  
\begin{bmatrix}
    \cos(\theta) \\
    \sin(\theta)
\end{bmatrix}.
\end{align}

\paragraph{Rotation symmetry detection.}
For rotation symmetry, our goal is to predict the positions of rotation centers and classify their fold (symmetry order). An $n$-fold rotational symmetry means the pattern looks the same after rotation by $\tfrac{2\pi}{n}$ (for example, a 4-fold symmetry repeats every $\tfrac{\pi}{2}$). To predict both axis existence and fold class, rotation branch $\mathcal{B}_\mathrm{rot}$ produces the multi-class classification score map:
\begin{align}
    \mathbf{O}_\mathrm{rot} = \mathcal{B}_\mathrm{rot}(\text{Pool}_G(\mathbf{F})) \in \mathbb{R}^{H \times W \times S},
\end{align}
where $S$ is the number of fold classes including the background class. Each rotation axis prediction is represented as $(x, y, p_s)$, with $p_s$ as the probability of the $s$-th fold class. 

\paragraph{Training objective.}
The training objective includes both reflection and rotation symmetry losses.
For reflection symmetry, we apply losses for midpoint classification, length regression, and orientation regression. 
Midpoint classification is optimized using weighted binary cross-entropy:
\begin{align}
\mathcal{L}_{p} = \mathbb{E}_{(x, y)} [ -\gamma_\mathrm{ref} p \log(\hat{p}) - (1 - p)\log(1 - \hat{p})],
\end{align}
where $p$ is the ground truth label, $\hat{p}$ is the predicted probability, and $\gamma_\mathrm{ref}$ is a weighting factor.
Length $\rho$ and orientation $\theta$ regression losses are applied only at positions with valid midpoints $(p = 1)$, enforced by the indicator function $\mathbb{I}_{p=1}$:
\begin{align}
\mathcal{L}_{\rho} &= \mathbb{E}_{(x, y)} [\mathbb{I}_{p=1} \cdot \mathrm{SmoothL1}(\rho, \hat{\rho})],  \\
\mathcal{L}_{\theta} &= \mathbb{E}_{(x, y)} [\mathbb{I}_{p=1} \cdot |\theta - \hat{\theta}|],
\end{align}
where $\rho$ and $\theta$ denote ground truth values, and $\hat{\rho}$ and $\hat{\theta}$ are the predicted values.
For rotation symmetry, fold classification is optimized using weighted multi-class cross-entropy:
\begin{align}
    \mathcal{L}_\mathrm{fold} = \mathbb{E}_{(x, y, s)}[-\gamma_\mathrm{rot} p_s \log \hat{p}_s],
\end{align}
where $s$ represents the fold class and $\gamma_\mathrm{rot}$ is applied at positions with ground truth rotation axis and correct fold class $\small{(p=1)}$. The total loss is a weighted sum of loss terms:
\begin{align}
\mathcal{L}_{\mathrm{total}} = \mathcal{L}_{p} + \lambda_{\rho} \mathcal{L}_{\rho} + \lambda_{\theta} \mathcal{L}_{\theta} + \lambda_\mathrm{fold} \mathcal{L}_\mathrm{fold}.
\end{align}

\captionsetup[figure]{skip=2pt}
\begin{figure}[t]
    \centering
    \includegraphics[width=\columnwidth]{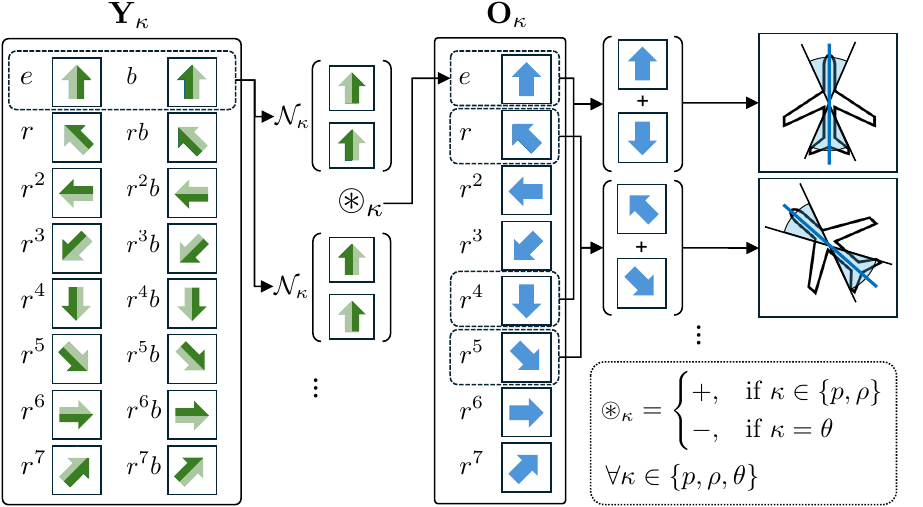} 
    \caption{\textbf{Illustration of our orientational anchor expansion on ${\mathrm{\mathbf D}_8}$ group.} The $\mathrm{D}_8$-equivariant features $\mathbf{Y}_\kappa$ undergo transformation $\mathcal{N}_\kappa$ and aggregation $\circledast_\kappa$, creating $\mathrm{C}_8$-equivariant features $\mathbf{O}_\kappa$. These are combined across opposite orientations to handle the $\theta$ and $\theta + \pi$ equivalence, allowing each orientation channel to specialize in specific angular ranges and improve detection of axes with overlapping midpoints. Each arrow represents a feature map.}
    \label{fig:orientational anchor expansion}
\end{figure}
\captionsetup[figure]{skip=2pt}
\begin{figure*}[t]
    \centering
    \includegraphics[width=\textwidth]{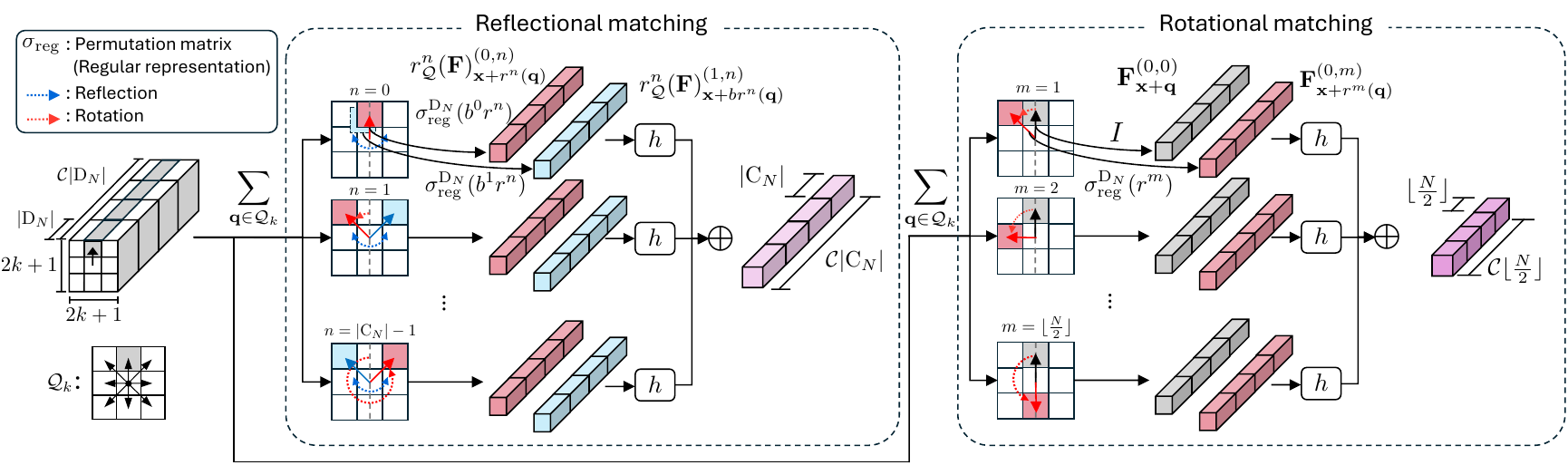}  
    \caption{\textbf{Illustration of our equivariant reflectional matching (left) and invariant rotational matching (right) modules.} The reflectional matching computes similarity scores between rotated features and their reflections across all $|\mathrm{C}_N|$ rotation angles, preserving dihedral group equivariance with rotation invariance. The rotational matching computes similarities between feature pairs with different rotation angle interval, yielding rotation-invariant features for detecting $n$-fold rotation symmetry centers. Both modules incorporate spatial neighborhoods $\mathcal{Q}_k$ for robust detection across multiple scales.}
    \label{fig:matching}
    \vspace{-4mm}
\end{figure*}
\subsection{Orientational anchor expansion}
\label{subsec:Orientational Anchor Expansion}
In standard object detection, anchor boxes~\cite{ren2015faster, liu2016ssd} are placed at various scales and aspect ratios to guide bounding box regression. Analogously, our reflection symmetry branch initially treats each pixel as an anchor for a potential symmetry axis and learns to regress the orientation and length of the axis. However, this straightforward approach does not fully exploit the orientation dimension provided by our group-equivariant features. To address this, we introduce orientational anchor expansion, integrating the group dimension into the detection framework for orientationally specialized axis detection and 
improved handling of axes with overlapping midpoints but different orientations. \Fig{orientational anchor expansion} illustrates the proposed orientational anchor expansion.
\vspace{-2mm}
\paragraph{Reflectional counterpart aggregation.}
In a $\mathrm{D}_N$-equivariant feature map, each group dimensional channel of $\mathbf{F}$ corresponds to a particular element of the dihedral group. Recall that reflection branch outputs in ~\Eq{reflection branch output} produce tensors $\mathbf{Y}_\kappa \in \mathbb{R}^{|\mathrm{D}_N| \times H \times W}$ for each component $\kappa \in \{p, \rho, \theta\}$. 
The $2N$ channels in this first dimension can be thought of as $N$ pairs, where each pair $(i,\ i+N)$ consists of a feature responding to some rotation $r^i$ and its reflected version $br^i$. To make use of the orientation-specific information, we aggregate each such pair of reflection counterparts into a single response, in a way that preserves the feature’s equivariance under pure rotations.

For the reflection symmetry midpoint scores $\mathbf{Y}_p$ and length $\mathbf{Y}_\rho$, which are unchanged by reflecting the image, we add the two responses. For the orientation output $\mathbf{Y}_\theta$, which flips sign under reflection (an axis at angle $\theta$ becomes $-\theta$), we subtract the reflected response. Formally, let $\mathbf{Y}_\kappa^{(i)}$ denote the feature map for the $i$-th rotation channel and $\mathbf{Y}_\kappa^{(i+N)}$ the corresponding reflection channel. We compute an aggregated feature $\tilde{\mathbf{Y}}_\kappa$ with only $N$ channels as:
\begin{equation}
\footnotesize{
\tilde{\mathbf{Y}}_\kappa = \bigoplus^{|\mathrm{C}_N|}_{i=1} \left[ \mathcal{N}_\kappa([{\mathbf{Y}}_\kappa^{(i)}; {\mathbf{Y}}_\kappa^{(i+N)}]) \circledast_\kappa \mathcal{N}_\kappa([{\mathbf{Y}}_\kappa^{(i+N)}; {\mathbf{Y}}_\kappa^{(i)}]) \right],}
\end{equation}
where $\circledast_\kappa$ is the pairwise aggregation operator defined as $\circledast_\kappa = +$ for $\kappa \in \{p, \rho\}$ and $\circledast_\kappa = -$ for $\kappa = \theta$. Rather than directly adding or subtracting feature maps which discards useful details, we apply a learnable transformation $\mathcal{N}_\kappa$ to extract and reweight information from each channel before combining, while preserving both the reflection transformation properties and $\mathrm{C}_N$-equivariance.

\paragraph{Orientational anchor construction.}
Even after merging reflection pairs, there remains an ambiguity in the orientation representation: a line at orientation $\theta$ is equivalent to the one at $\theta + \pi$, since both describe the same physical axis line. To address this ambiguity, we combine the aggregated response at rotation channel index $\alpha$ with that at $\alpha + \sfrac{N}{2}$ to produce $\mathbf{O}_\text{ref} =[\mathbf{O}_p; \mathbf{O}_\rho; \mathbf{O}_\theta]\in \mathbb{R}^{{\sfrac{|\mathrm{C}_N|}{2}} \times H \times W \times 3 }$ :
\begin{equation}
\mathbf{O}_{\kappa, \alpha} = \tilde{\mathbf{Y}}_{\kappa, \alpha} + \tilde{\mathbf{Y}}_{\kappa, \alpha + \sfrac{N}{2}}, \quad \alpha = 1, \dots, \tfrac{N}{2},
\end{equation}
for each component $\kappa \in \{p, \rho, \theta\}$. Each anchor $\mathbf{O}_\alpha$ specializes in detecting axes with orientation offsets within $[-{\tfrac{\pi}{N}}, {\tfrac{\pi}{N}})$ from its base orientation $\tfrac{2\pi\alpha}{N}$. We predict offsets rather than absolute orientations to directly adapt invariant orientation regression values across different anchor orientations. At each position $(\alpha, x, y)$, an axis is represented as $(\alpha, x, y, p, \rho, \theta)$, where the output $\mathbf{O}_{(\alpha, x, y)} = (p, \rho, \theta)$ determines its start and end points as:
\begin{align}
\begin{bmatrix}
    x_{\text s, \alpha}\\
    y_{\text s, \alpha}
\end{bmatrix}
&=
\begin{bmatrix}
    x_\alpha \\
    y_\alpha
\end{bmatrix}
+ \frac{\rho}{2} \begin{bmatrix}
    \cos(\theta_\alpha) \\
    \sin(\theta_\alpha)
\end{bmatrix}, \\
\begin{bmatrix}
    x_{\text e, \alpha} \\
    y_{\text e, \alpha}
\end{bmatrix}
&=
\begin{bmatrix}
    x_\alpha \\
    y_\alpha
\end{bmatrix}
- \frac{\rho}{2}  \begin{bmatrix}
    \cos(\theta_\alpha) \\
    \sin(\theta_\alpha)
\end{bmatrix},
\end{align}
where $\theta_\alpha = \tfrac{2\pi\alpha}{N} + \theta$ represents the absolute orientation.

\subsection{Reflectional matching}
\label{subsec:Reflectional Matching}
Reflection symmetry can be validated by comparing a pattern with its mirrored counterpart, known as reflectional matching~\cite{loy2006detecting, cho2009bilateral}. Unlike hand-crafted descriptors such as SIFT~\cite{lowe2004distinctive}, conventional neural features~\cite{he2016deep, dosovitskiy2020image} lack rotation and reflection equivariance, limiting their effectiveness. To address this, we leverage $\mathrm{D}_N$-equivariant features~\cite{cohen2016group} for reflectional matching, providing a strong cue for symmetry detection. \Fig{matching} (left) illustrates the detailed process.

For a feature fiber $\mathbf{f} \in \mathbb{R}^{\mathcal{C} |\mathrm{D}_N|}$ from a $\mathrm{D}_N$-equivariant feature map $\mathbf{F}$, its transformation under $l$ reflections and $n$ rotations is:
\begin{align}
    \mathbf{f}^{(l,n)} =  \bigoplus_{c=1}^{\mathcal{C}} \sigma_{\mathrm{reg}}^{\mathrm{D}_N}(b^l r^n) \mathbf{f}_c \in \mathbb{R}^{\mathcal{C}|\mathrm{D}_N|},
\end{align}
where $\mathbf{f}_c \in \mathbb{R}^{|\mathrm{D}_N|}$ represents the group-equivariant subset of the fiber, and $\mathbf{f}=[\mathbf{f}_1^\top, \dots, \mathbf{f}_\mathcal{C}^\top]^\top$. Here, $\sigma_{\mathrm{reg}}^{\mathrm{D}_N}(b^l r^n)$ denotes the regular representation of $\mathrm{D}_N$ for $l$ reflections and $n$ rotations.
The group-aware similarity $h$ between two fibers $\mathbf{f}^1, \mathbf{f}^2 \in \mathbb{R}^{\mathcal{C} |\mathrm{D}_N|}$ is defined as:
\begin{align}
    h(\mathbf{f}^1, \mathbf{f}^2) = \bigoplus_{c=1}^\mathcal{C} \frac{\mathbf{f}^1_c \cdot \mathbf{f}^2_c}{\| \mathbf{f}^1_c \| \| \mathbf{f}^2_c \|} \in \mathbb{R}^\mathcal{C}.
\end{align}
To capture symmetry across orientations, reflectional similarity scores are computed for each rotation, comparing rotated and rotated-then-reflected fibers:
\begin{align}
    \mathbf{H}_{\text{ref}, \mathbf{x}} = \bigoplus_{n=0}^{|\mathrm{C}_N|-1} h(\mathbf{F}^{(0,n)}_\mathbf{x}, \mathbf{F}^{(1,n)}_\mathbf{x})\in\mathbb{R}^{\mathcal{C}|\mathrm{C}_N|},
    \label{eq:single-erm}
\end{align}
where $\mathbf{F}_{\mathbf{x}}^{(0,n)}$ and $\mathbf{F}_{\mathbf{x}}^{(1,n)}$ represent fibers at position $\mathbf{x}$ under the regular representation for $n$ rotations, with and without reflection. The resulting similarity score map $\mathbf{H} \in \mathbb{R}^{\mathcal{C} |\mathrm{C}_N|\times H \times W}$ is equivariant under the dihedral group while remaining reflection-invariant.
To detect broader symmetries beyond single points, we extend matching to spatial neighborhoods defined by a set of 2D offset vectors:
\begin{align}
  &\mathcal{Q}_k = \left\{ (i, j) \mid i, j \in \{-k, \dots, k \} \right\},
\end{align}
where $k\in\mathbb{N}$ controls the neighborhood size. The neighborhood similarity is computed as:
\begin{equation}
    \footnotesize
    {\mathbf{H}^{(k)}_{\text{ref},\mathbf{x}} = \sum_{\mathbf{q} \in \mathcal{Q}_k}  \bigoplus_{n=0}^{|\mathrm{C}_N|-1} h(r^n_\mathcal{Q}(\mathbf{F})^{(0,n)}_{\mathbf{x} + {r^n}(\mathbf{q})}, r^n_\mathcal{Q}(\mathbf{F})^{(1,n)}_{\mathbf{x} + {b r^n}(\mathbf{q})})\in\mathbb{R}^{\mathcal{C}|\mathrm{C}_N|}},
    \label{eq:spatial-erm}
\end{equation}
where $b^l r^n(\mathbf{q})$ denotes the transformed offset after $n$ rotations and $l$ reflections and $r^n_\mathcal{Q}$ denotes the rotation of the entire neighborhood $\mathcal{Q}$ about its center. 
To improve robustness, we use multi-scale reflectional similarity features $\mathbf{H}_\text{ref}^{(k_1)}, \dots, \mathbf{H}_\text{ref}^{(k_M)}$, concatenated with the base feature map $\mathbf{F}$ to capture symmetry across various spatial scales while preserving equivariance. The matching output is equivariant to $\mathrm{D}_N$ while preserving reflection invariance, as demonstrated in the detailed proof provided in the Appendix~\ref{appendix B}. 

\subsection{Rotational matching}
\label{subsec:Rotational Matching}
Rotation symmetry is identified by comparing a pattern with its rotated version around its axis. Our rotational matching module implements this by comparing features with their rotated versions around each candidate center point (\Fig{matching} right). An $n$-fold rotational symmetry remains invariant under every $\frac{2\pi}{n}$ rotation. To reduce redundancy in similarity comparisons, we exploit the consistency of feature comparisons at fixed angular separations, requiring only $\lfloor\frac{N}{2}\rfloor$ unique comparisons instead of $_N\text{C}_2$ feature pairs. The complete rotational matching feature is computed as:
\begin{align}
    \mathbf{H}_\text{rot,\textbf{x}} = \bigoplus_{m=1}^{\lfloor\frac{N}{2}\rfloor} h(\mathbf{F}^{(0,0)}_\mathbf{x}, \mathbf{F}^{(0,m)}_\mathbf{x}) \in \mathbb{R}^{\mathcal{C}\lfloor\frac{N}{2}\rfloor},
\end{align}
which remains dihedral group-invariant, as similarity values are preserved. To extend matching to spatial neighborhoods, we use the approach from \Subsec{Reflectional Matching}:  
\begin{align}
    \mathbf{H}^{(k)}_\text{rot,\textbf{x}} = \sum_{\mathbf{q} \in \mathcal{Q}_k} \bigoplus_{m=1}^{\lfloor\frac{N}{2}\rfloor} h(\mathbf{F}^{(0,0)}_{\mathbf{x} + \mathbf{q}}, \mathbf{F}^{(0,m)}_{\mathbf{x} + {r^m}(\mathbf{q})}) \in \mathbb{R}^{\mathcal{C}\lfloor\frac{N}{2}\rfloor}.
\end{align}
Following the multi-scale approach in \Subsec{Reflectional Matching}, we compute features at multiple kernel sizes and concatenate them with the pooled base feature map from \Subsec{Instance-level Symmetry Detection}, enabling precise detection of rotation axis and fold classes. The resulting output remains invariant to both rotation and reflection.

\begin{figure*}[t]
    \centering
    \newcommand{\imgwidtha}{0.125\textwidth}
    \newcommand{\imgwidthb}{0.125\textwidth}
    \setlength{\tabcolsep}{0.9pt}
    \resizebox{\textwidth}{!}{
    \begin{tabular}{cccccccc}
        \scriptsize PMCNet\citep{seoshim2021pmcnet} & \scriptsize EquiSym\citep{seo2022equisym} & \scriptsize Ours & \scriptsize Ground truth &
        \scriptsize PMCNet\citep{seoshim2021pmcnet} & \scriptsize EquiSym\citep{seo2022equisym} & \scriptsize Ours & \scriptsize Ground truth\\
        \includegraphics[width=\imgwidtha]{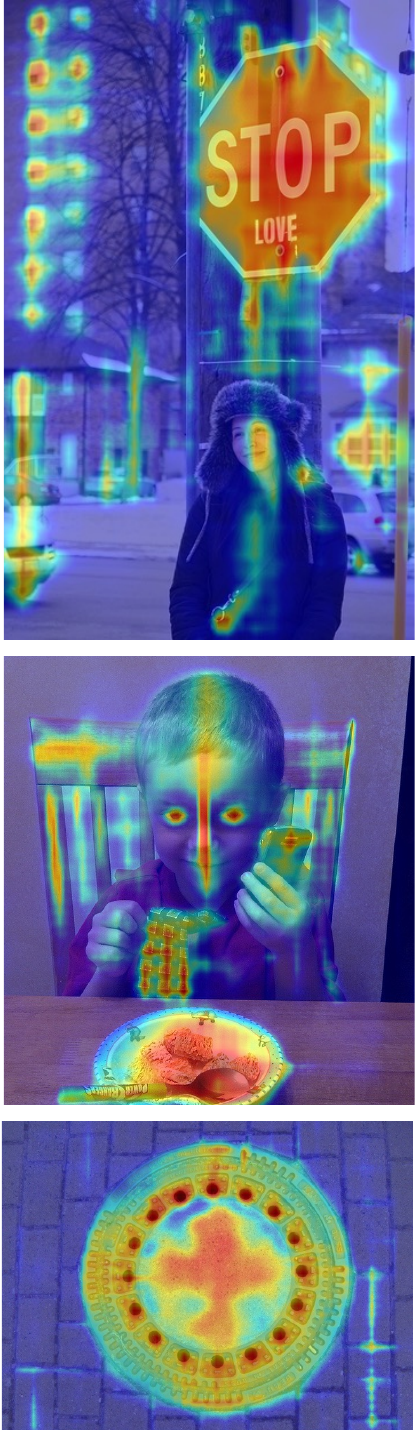} &
        \includegraphics[width=\imgwidtha]{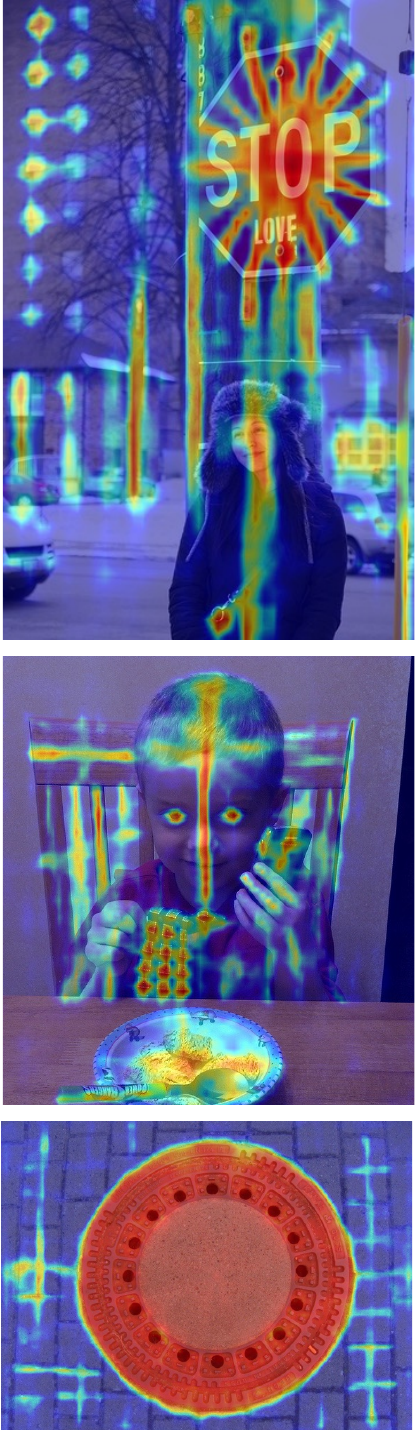} &
        \includegraphics[width=\imgwidtha]{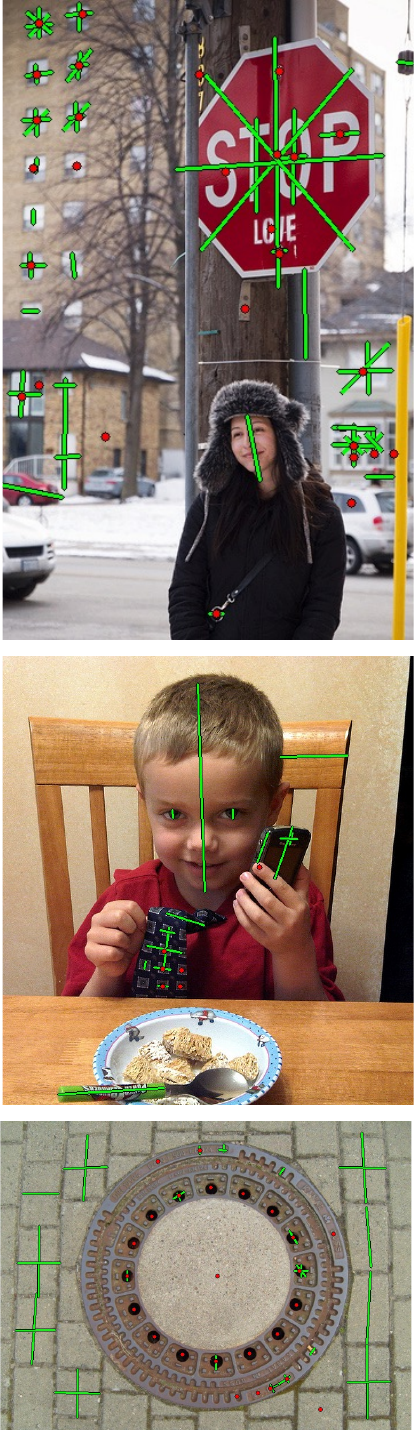} &
        \includegraphics[width=\imgwidtha]{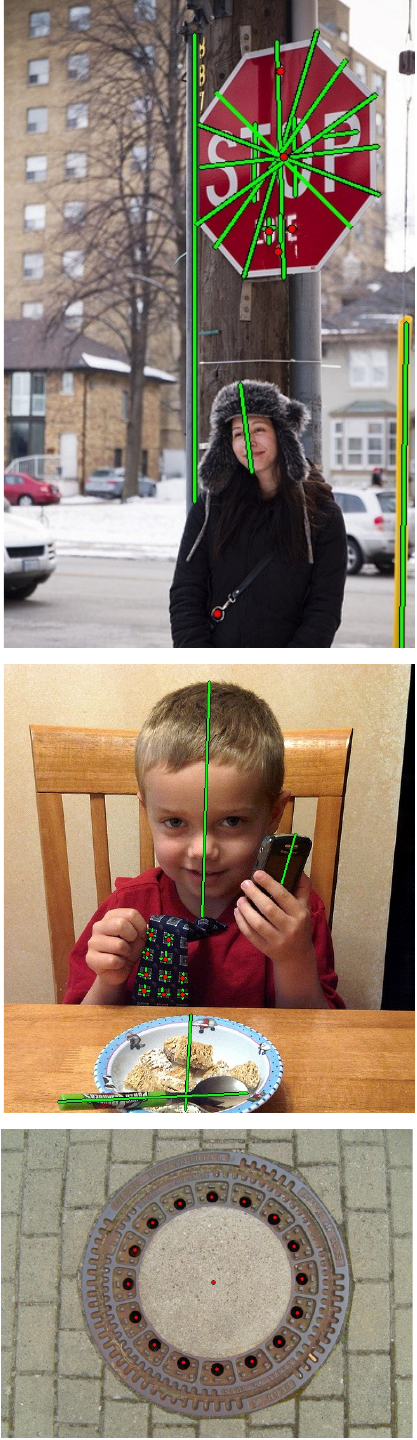} &
        \includegraphics[width=\imgwidthb]{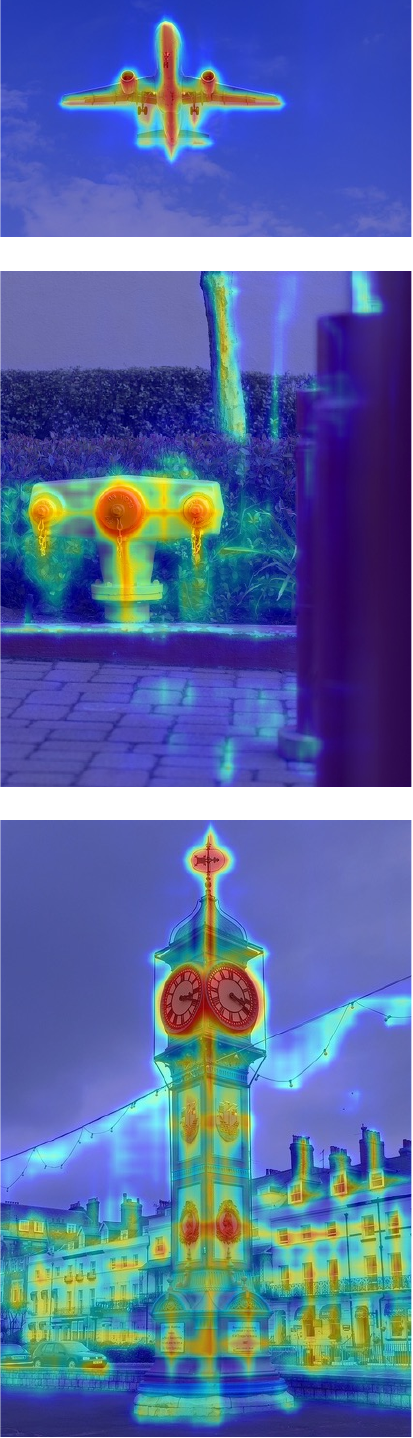} &
        \includegraphics[width=\imgwidthb]{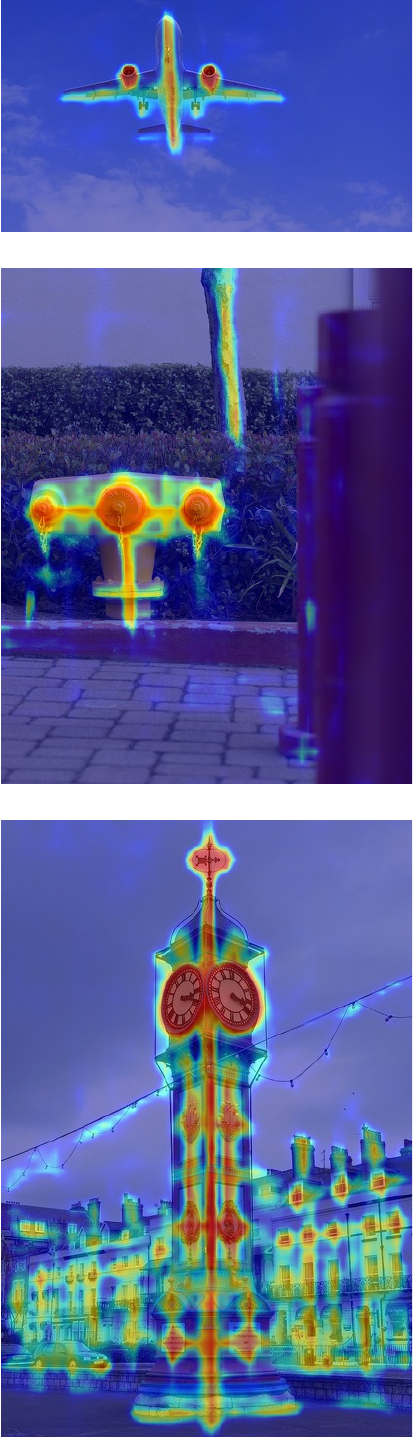} &
        \includegraphics[width=\imgwidthb]{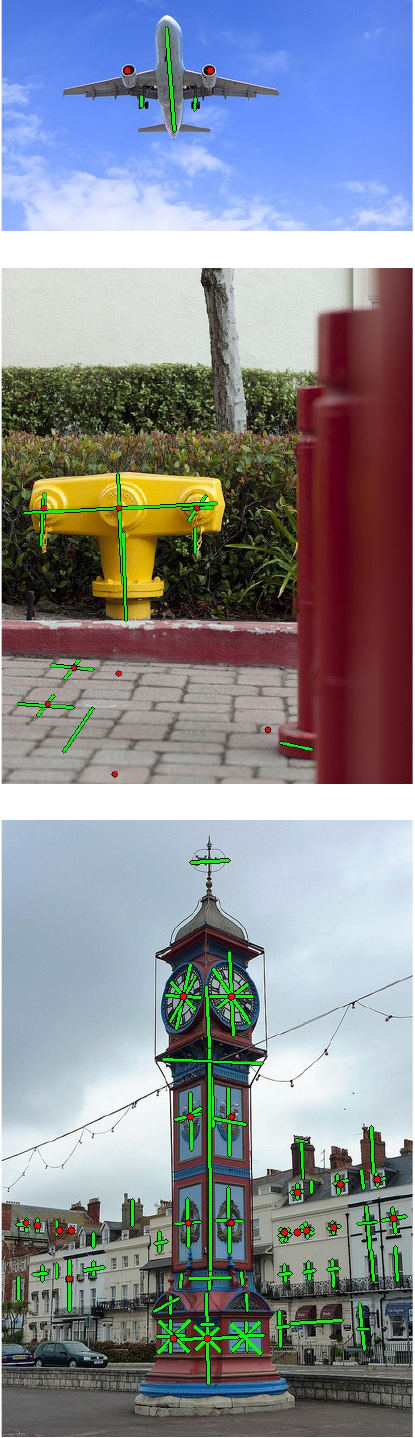} &
        \includegraphics[width=\imgwidthb]{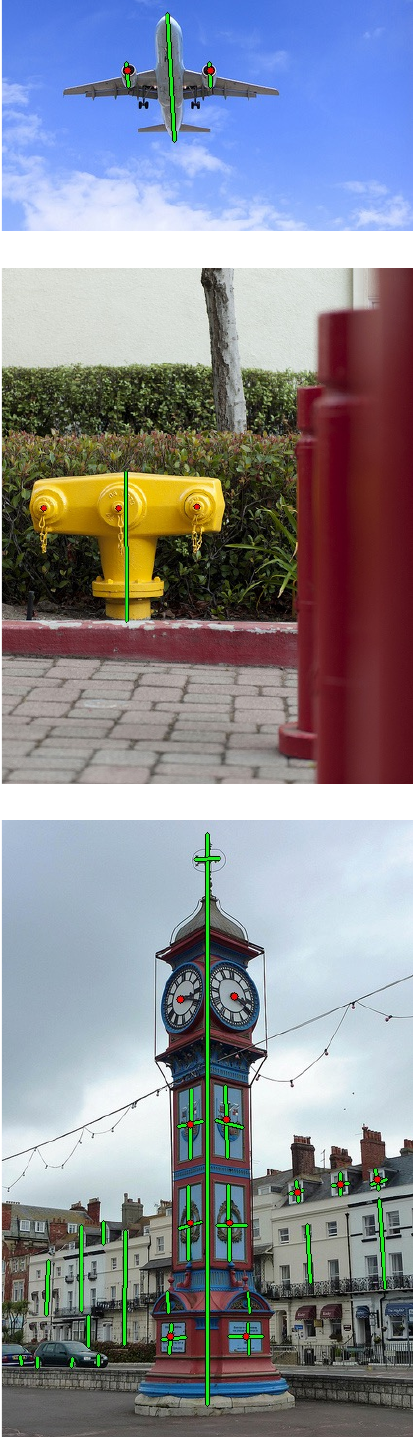} 
    \end{tabular}
    }
    \caption{\textbf{Qualitative comparison of symmetry detection methods.} Our instance-wise approach produces clearer, more precise symmetry instances compared to heatmap-based methods~\citep{seoshim2021pmcnet, seo2022equisym}, especially for smaller objects and complex scenes. Green lines in ground truth and our results represent reflection axes, while red points represent rotation axes.}
    \label{fig:qualitative_results}
    \vspace{-4mm}
\end{figure*}
\section{Experiments}
\subsection{Implementation details}
\paragraph{Dataset.}
We use the DENDI dataset~\cite{seo2022equisym} for both training and evaluation. It provides annotations for reflection symmetry axes and rotation symmetry centers, along with corresponding fold numbers. The original training set comprises approximately 1.8k images. To augment the data, we extract 15k symmetry-annotated masks, which are pasted onto other images without overlapping existing annotations, thereby expanding the dataset to 30k training images. We also report F1-scores on the SDRW~\cite{liu2013symmetry} and LDRS~\cite{seo2022equisym} datasets for comparison with prior methods.
\paragraph{Evaluation metrics.}
For reflection symmetry, we adopt structural Average Precision (sAP)~\cite{zhou2019end}, where a prediction is correct if $d_1^2 + d_2^2 < \tau$ or $d_p^2 < \tfrac{\tau}{2}$, with at least 70\% overlap within the annotated ellipse. Here, $d_1$ and $d_2$ are endpoint distances to the ground truth, and $d_\mu$ is the distance from the ellipse center to the predicted midpoint. For rotation symmetry, sAP is computed with $d_\text{center}^2 < \tfrac{\tau}{2}$, where $d_\text{center}$ denotes the distance between predicted and ground truth rotation centers. We also report fold sAP, which requires correct fold classification. All results are evaluated at $\tau = 5$, 10, and 15 pixels. For heatmap-based baselines~\cite{funk2017beyond, seo2022equisym}, F1-scores are computed using 5-pixel dilated ground-truth and predicted axes.

\paragraph{Model and training.}  
We use a $\mathrm{D}_8$-equivariant ResNet-34~\cite{he2016deep, cohen2016group} as the feature extractor. Both reflectional and rotational matching modules utilize multi-scale similarity features (scales 1, 3, and 5). In the reflection branch, group convolution~\cite{cohen2016group} is implemented via image rotation, group channel permutation, and standard convolution, as the \texttt{e2cnn}~\cite{e2cnn} framework does not support reflection-invariant dihedral groups or operations such as deformable convolution~\cite{dai2017deformable} while preserving equivariance. The model is trained for 100 epochs with a batch size of 32 using AdamW~\cite{Adamsolver}, starting with a learning rate of $10^{-3}$, reduced by a factor of 10 at epochs 50 and 75. Loss weights are set to $\lambda_{\rho} = 1$ (length), $\lambda_{\theta} = 150$ (orientation, accounting for radian scale), and $\lambda_\mathrm{fold} = 2$ (fold classification). Weighted binary cross-entropy with a positive class weight of 3 is used for $\mathcal{L}_{\mathrm{mid}}$ and $\mathcal{L}_{\mathrm{fold}}$.

\subsection{Evaluation of the proposed method}
\paragraph{Reflection symmetry detection.}  
As shown in the last row of the~\Tbl{ref_abl}, the proposed model achieves sAP scores of 18.7, 22.7, and 24.7 at 5, 10, and 15-pixel thresholds on the DENDI dataset~\cite{seo2022equisym}. \Fig{qualitative_results} demonstrates robust reflection symmetry detection across diverse scenes, handling multiple orientations and scales, even in complex backgrounds.
\paragraph{Rotation symmetry detection.}  
The rotation symmetry branch outputs classification scores for multiple folds. For center detection, we pool the maximum score as the center probability for binary evaluation. In the last row of the ~\Tbl{rot_abl}, we report both center sAP and fold sAP. Our method achieves center sAP scores of 36.8, 39.1, and 40.0 and fold AP scores of 26.6, 28.3, and 28.9 at 5, 10, and 15-pixel thresholds, respectively.  
Fold misclassifications primarily occur between 2-fold and 4-fold symmetries. \Fig{qualitative_results} shows robust detection across complex scenes.

\definecolor{lgray}{rgb}{0.9, 0.9, 0.9}
\begin{table}[t]
\centering
\fontsize{9}{10}\selectfont
\begin{tabular}{lccc}
\toprule
\multirow{2}{*}{Method} & \multicolumn{3}{c}{Ref. sAP (\%)} \\
 & @5 & @10 & @15\\
\midrule
Axis-level detection & 6.2 & 9.3 & 11.2 \\
\ \ + Orientational anchors & 16.6 & 19.9 & 21.1 \\
\ \ \ \ + Ref. match$_{k=0}$ & 17.6 & 20.7 & 21.8 \\
\ \ \ \ + Ref. match$_{k=0,1}$ & 18.4 & 22.0 & 23.7 \\
\rowcolor{lgray}
\ \ \ \ + Ref. match$_{k=0,1,2}$ & \textbf{18.8} & \textbf{22.7} & \textbf{24.7} \\
\bottomrule
\end{tabular}
\vspace{-1mm}
\caption{Ablation results for reflection symmetry detection on the DENDI dataset. Ref. match$_k$ represents reflectional matching with kernel sizes $2k+1$. Best results are shown in \textbf{bold}.}
\label{tab:ref_abl}
\vspace{-4mm}
\end{table}

\begin{figure*}[t]
    \centering
    \setlength{\tabcolsep}{0.9pt}
    
    \begin{minipage}[t]{0.7\textwidth}  
        \centering
        \begin{tabular}{cccc}
            \ \ \ \ \ \ \ \small{SDRW}\footnotesize{~\cite{liu2013symmetry} (Ref.)} & 
            \ \ \ \ \ \small{LDRS}\footnotesize{~\cite{seoshim2021pmcnet} (Ref.)} &  
            \ \ \ \ \ \small{DENDI}\footnotesize{~\cite{seo2022equisym} (Ref.)} &  
            \ \ \ \ \ \small{DENDI}\footnotesize{~\cite{seo2022equisym} (Rot.)} \\
            \includegraphics[width=0.24\textwidth]{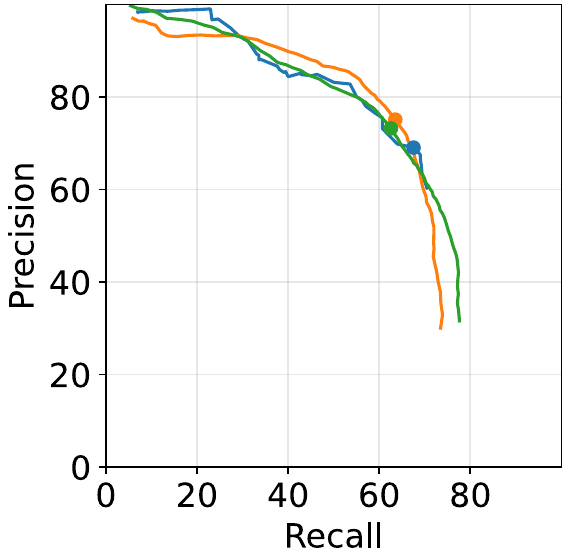} &
            \includegraphics[width=0.24\textwidth]{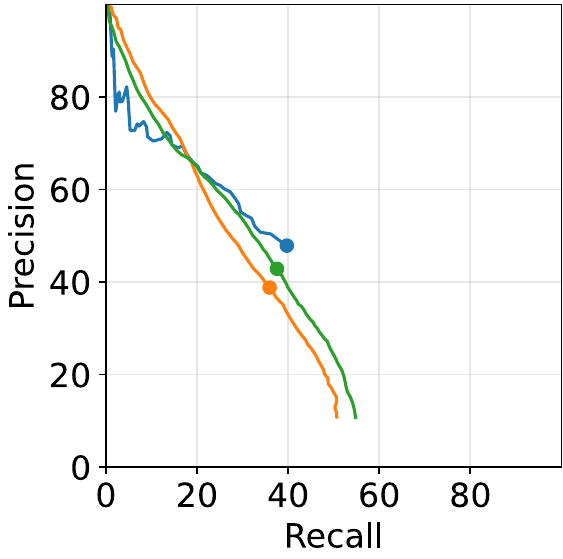} &
            \includegraphics[width=0.24\textwidth]{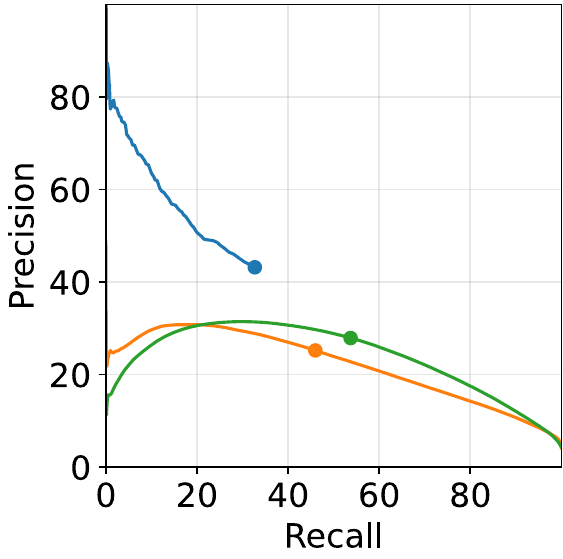} &
            \includegraphics[width=0.24\textwidth]{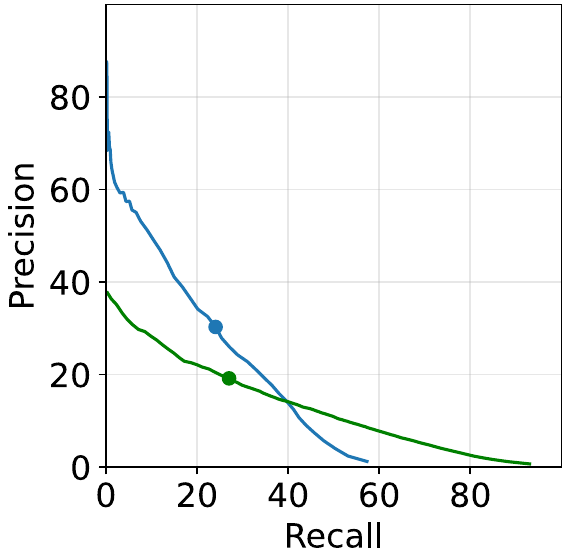} \\
            \multicolumn{4}{c}{\footnotesize{(a)}}\\
        \end{tabular}
        \label{fig:pr_curves}
    \end{minipage}
    \hfill
    \begin{minipage}[t]{0.18\textwidth} 
        \centering
        \begin{tabular}{c}
        \ \ \ \small{SDRW}\footnotesize{~\cite{liu2013symmetry}} \small{(Padding vs F1)} \\
        \includegraphics[width=0.95\textwidth]{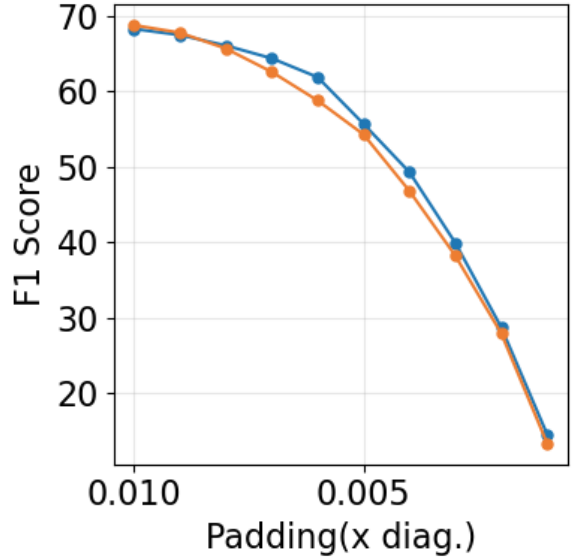}   \\
        \ \ \ \footnotesize{(b)}\\
        \end{tabular}
        \label{fig:f1_padding}
    \end{minipage}
    \hfill
    \begin{minipage}[t]{0.08\textwidth} 
        \centering
        \begin{tabular}{c}
        \includegraphics[width=\textwidth]{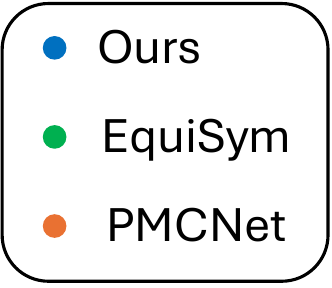}\\
        \end{tabular}
        \label{fig:f1_padding_legend}
    \end{minipage}
    \vspace{-1mm}
    \caption{\textbf{Evaluation of our symmetry detection approach.} (a) Precision-recall curves for reflection symmetry detection on SDRW~\cite{liu2013symmetry}, LDRS~\cite{seoshim2021pmcnet}, and DENDI~\cite{seo2022equisym} datasets, and rotation symmetry detection on the DENDI dataset; (b) Analysis of F1-score with varying axis padding values(multiplied to image diagonal) on the SDRW dataset. For rotation symmetry evaluation on DENDI, we compare only with EquiSym~\cite{seo2022equisym}, as PMCNet~\cite{seoshim2021pmcnet} does not support rotation symmetry detection. }
    \label{fig:prcurves}
    \vspace{-3mm}
\end{figure*}
\subsection{Ablation study}
\paragraph{Reflection symmetry detection.}
We conduct ablation studies on our reflection symmetry components in~\Tbl{ref_abl}. The baseline axis-level detection achieves sAP scores of 6.2, 9.3, and 11.2. Adding orientational anchors significantly improves performance, increasing sAP to 16.6, 19.9, and 21.1 by enabling orientation-specific feature learning.  
Incorporating single-kernel reflectional matching ($k\mspace{-3mu}=\mspace{-3mu}0$) further boosts performance, achieving sAP scores of 17.6, 20.7, and 21.8. Expanding to multi-kernel matching ($k\mspace{-3mu}=\mspace{-3mu}0,1$) enhances detection, reaching 18.4, 22.0, and 23.7. The best performance is obtained with $k\mspace{-3mu}=\mspace{-3mu}0,1,2$, achieving sAP scores of 18.8, 22.7, and 24.7. These results confirm that multi-scale reflectional matching effectively captures symmetry patterns across different spatial scales.

\definecolor{lgray}{rgb}{0.9, 0.9, 0.9}
\begin{table}[t]
\centering
\fontsize{9}{10}\selectfont
{
\begin{tabular}{lccc}
\toprule
\multirow{2}{*}{Method} & \multicolumn{3}{c}{Center sAP \footnotesize{(Fold sAP)} (\%)}\\
& @5 & @10 & @15 \\
\midrule
Axis-level detection & 31.5\footnotesize{(22.5)} & 34.7\footnotesize{(24.6)} & 35.7\footnotesize{(25.3)} \\
\ + Rot. match$_{k=0}$ & 35.9\footnotesize{(25.4)} & 37.8\footnotesize{(26.6)} & 37.0\footnotesize{(27.2)} \\
\ + Rot. match$_{k=0,1}$ & 36.2\footnotesize{(26.2)} & 38.2\footnotesize{(27.8)} & 37.4\footnotesize{(28.1)} \\
\rowcolor{lgray}
\ + Rot. match$_{k=0,1,2}$ & \textbf{36.8}\footnotesize{(\textbf{26.6})} & \textbf{39.1}\footnotesize{(\textbf{28.3})} & \textbf{40.0}\footnotesize{(\textbf{28.9})} \\
\bottomrule
\end{tabular}
}
\vspace{-2mm}
\caption{Ablation results for rotation symmetry detection on the DENDI dataset. Rot. match$_k$ represents rotational matching with kernel sizes $2k+1$. Best results are shown in \textbf{bold}.}
\label{tab:rot_abl}
\vspace{-2mm}
\end{table}
\paragraph{Rotation symmetry detection.}
\Tbl{rot_abl} shows the effectiveness of our rotational matching approach. The baseline axis-level detection achieves sAP scores of 31.5, 34.7, and 35.7, with fold sAP scores of 22.5, 24.6, and 25.3. Adding single-kernel rotational matching ($k\!=\!0$) improves sAP by 4.4, 3.1, and 1.3, with greater gains at smaller thresholds (5 pixels), indicating better localization.  
Expanding to multi-kernel matching ($k\!=\!0,1$ and $k\!=\!0,1,2$) further enhances performance. Our final model achieves sAP scores of 36.8, 39.1, and 40.0, with fold sAP scores of 26.6, 28.3, and 28.9, demonstrating the effectiveness of rotational matching across scales.

\begin{figure}[h]
    \centering
    \includegraphics[width=\linewidth]{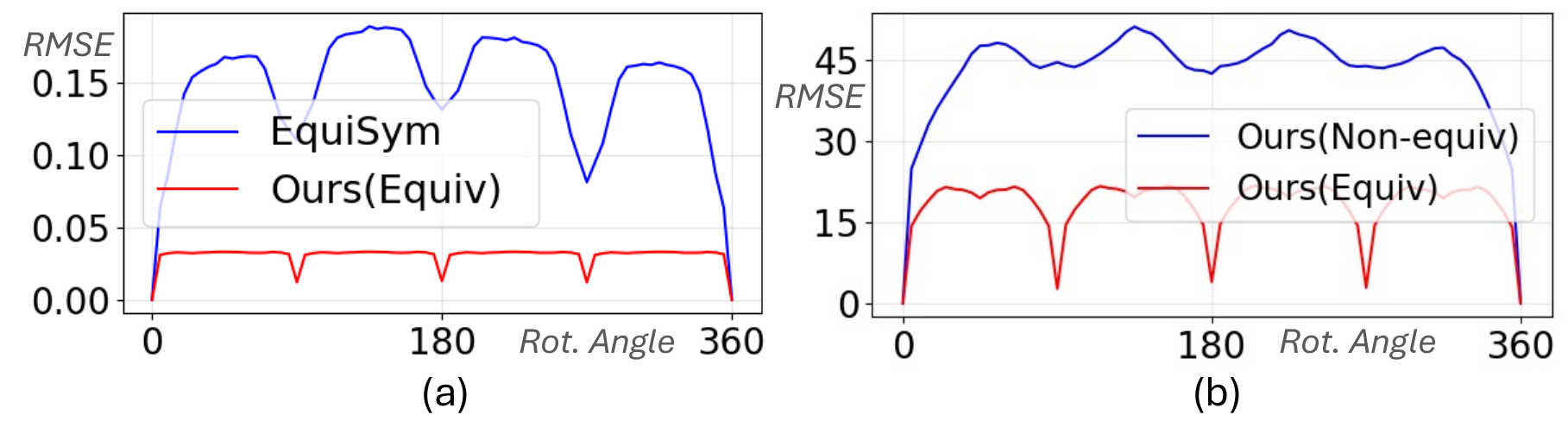}
    \caption{\textbf{Rotation robustness evaluation on the DENDI test set.} (a) Comparison with the heatmap-based equivariant method~\cite{seo2022equisym}. (b) Comparison of $\mathbf{O}_\rho$ between equivariant and non-equivariant variants of our model.}
    \label{fig:rot_rmse_main}
    \vspace{-4mm}
\end{figure}

\paragraph{Equivariance Analysis} 
To quantitatively validate the rotation equivariance illustrated in \Fig{teaser}, we compute the RMSE between outputs from non-rotated and rotated inputs on the DENDI test set. In Fig.~\ref{fig:rot_rmse_main} (a), axis-level predictions are converted to heatmaps for fair comparison with the heatmap-based group-equivariant method~\cite{seo2022equisym}. Our model consistently achieves lower RMSE across all rotation angles, demonstrating greater robustness. In Fig.~\ref{fig:rot_rmse_main} (b), we compare the length map $\mathbf{O}_\rho$ between the equivariant and non-equivariant variants of our model. The equivariant version yields significantly lower RMSE, confirming the effectiveness of our equivariant design in preserving rotation robustness.
\begin{table}
\centering
\fontsize{9}{10}\selectfont
\scalebox{0.95}{
\begin{tabular}{@{}l@{\hspace{6pt}}c@{\hspace{6pt}}c@{\hspace{6pt}}c@{\hspace{6pt}}c@{}}
\toprule
\multirow{2}{*}{Method} & \multicolumn{3}{c}{\small{Ref. F1 (\%)}} & \small{Rot. F1 (\%)} \\
\cmidrule(lr){2-4} \cmidrule(lr){5-5}
 & \small{SDRW}\footnotesize{~\cite{liu2013symmetry}} & \small{LDRS}\footnotesize{~\cite{seoshim2021pmcnet}} & \small{DENDI}\footnotesize{~\cite{seo2022equisym}} & \small{DENDI}\footnotesize{~\cite{seo2022equisym}} \\
\midrule
\small{PMCNet}\footnotesize{~\cite{seoshim2021pmcnet}} & \textbf{68.8} & 37.3 & 32.6 & - \\
\small{EquiSym}\footnotesize{~\cite{seo2022equisym}} & 67.5 & 40.0 & 36.7 & 22.4 \\
Ours & 68.3 & \textbf{43.4} & \textbf{37.2} & \textbf{26.8} \\
\bottomrule
\end{tabular}
}
\vspace{-2mm}
\caption{Comparison with the state-of-the-art methods using pixel-wise F1-score on multiple datasets. Ref. and Rot. denote reflection and rotation symmetry respectively. Best results in \textbf{bold}.}
\label{tab:f1_refrot_dendi}
\vspace{-3mm}
\end{table}

\subsection{Comparison with the state-of-the-art methods}
\label{subsec:comparison with sota}
\paragraph{F1-score.}
\Tbl{f1_refrot_dendi} shows F1-scores across multiple benchmarks. Our method outperforms previous work on LDRS~\cite{seoshim2021pmcnet} (+3.4) and DENDI~\cite{seo2022equisym} (+0.5) for reflection symmetry and achieves a significant gain(+4.4) in rotation symmetry detection on DENDI. For the SDRW~\cite{liu2013symmetry}, our method (68.3) is comparable to PMCNet (68.8). This slight difference stems from the disparity between heatmap-based segmentation and detection approaches. As shown in \Fig{prcurves}(b), when the evaluation criterion becomes more stringent (smaller padding values), our axis-level approach outperforms PMCNet's region-based predictions due to more precise axis localization.

\paragraph{PR curve.} 
Precision-recall curves in \Fig{prcurves}(a) further highlight performance differences. Our method maintains higher precision, especially in LDRS and DENDI. Unlike pixel-level methods that boost recall by predicting all pixels, our approach models symmetry as geometric primitives, where a single midpoint score affects the entire axis. This results in higher precision but more variable recall. Post-processing steps like Non-Maximum Suppression and score thresholding further prioritize precision over recall.

\section{Conclusion}
We have introduced a dihedral group-equivariant approach for axis-level symmetry detection, representing symmetries as geometric primitives instead of pixel-level heatmaps. Our method integrates orientational anchors and reflectional matching for reflection symmetry detection, and invariant rotational matching for rotation symmetry detection to capture symmetry across orientations and scales. 
Experiments demonstrate superior performance over existing methods, with ablations validating the effectiveness of our approach. Future work can extend our model to continuous groups, 3D spaces, and varying viewpoints for real-world applications.

\section*{Acknowledgements}

This work was supported by IITP grants (RS-2022-II220290: Visual Intelligence for Space-Time Understanding \& Generation (60\%), RS-2024-00457882: National AI Research Lab Project (35\%), RS-2019-II191906: AI Graduate School Program at POSTECH (5\%)) funded by Ministry of Science and ICT, Korea.

{
    \small
    \bibliographystyle{ieeenat_fullname}
    \bibliography{main}
}

\clearpage
 
\renewcommand{\thesection}{\Alph{section}}

\appendix

\setcounter{table}{0}

\setcounter{figure}{0}

\setcounter{page}{1}
\setcounter{equation}{0}

\maketitlesupplementary
 
\renewcommand{\thefigure}{A\arabic{figure}}

\renewcommand{\thetable}{A\arabic{table}}

\section{Regular representation and group convolution}
\addcontentsline{toc}{section}{Appendix A: Regular representation and group convolution}
\label{appendix A}
\subsection{Discrete group representation}
\paragraph{Regular group representation.}
The regular representation of a finite group $G = \{g_1, \dots, g_N\}$ acts on a vector space $\mathbb{R}^{|G|}$. For any element $g \in G$, the regular representation $\sigma^{G}_{\mathrm{reg}}(g)$ is defined as:
\begin{align}
    \sigma^{G}_{\mathrm{reg}}(g) = [\mathbf{e}_{g \cdot g_1}, \dots, \mathbf{e}_{g \cdot g_N}],
\end{align}
where each group element $g_i \in G$ is associated with a basis vector $\mathbf{e}_{g_i} \in \mathbb{R}^{|G|}$. In regular representation, $\sigma_\mathrm{reg}^G(g) \in \mathbb{R}^{|G| \times |G|}$ is a permutation matrix that maps each basis vector $\mathbf{e}_{g_i}$ to $\mathbf{e}_{g \cdot g_i}$ for all $g_i \in G$.
\paragraph{Cyclic group representation.}
The cyclic group $\mathrm{C}_N$, consisting of $N$ discrete planar rotations, is defined as $\{r^0, r^1, \dots, r^{(N-1)}\}$ with rotation generator $r$. With the group law $r^a \cdot r^b = r^{(a+b) \bmod N}$, the regular representation of $r^n$ is given by:
\begin{align}
    \sigma^{\mathrm{C}_N}_{\mathrm{reg}}(r^n) 
    &= [\mathbf{e}_{r^n}, \mathbf{e}_{r^{(n+1) \bmod N}}, \dots, \mathbf{e}_{r^{(n+N-1) \bmod N}}],
\end{align}
where the basis vectors are defined from:
\begin{align}
    \sigma_\mathrm{reg}^{\mathrm{C}_N}(r^0) = \mathrm{I}_N,
\end{align}
where $\mathrm{I}_N$ being the $N\times N$ identity matrix. Here, the regular representation of the cyclic group corresponds to a cyclic permutation matrix.
\paragraph{Dihedral group representation.}
The dihedral group $\mathrm{D}_N = \{r^0, \dots, r^{N-1}, b, rb, \dots, r^{N-1}b\}$, consisting of $2N$ elements, is an extension of the cyclic group that includes an additional reflection generator $b$. The regular representation of the element $r^n b$ is given by: 
\begin{align}
\sigma^{\mathrm{D}_N}_{\mathrm{reg}}(r^n b) 
    &= [\mathbf{e}_{r^n b}, \mathbf{e}_{r^n b \cdot r}, \dots, \mathbf{e}_{r^n b \cdot r^{N-1}},\nonumber \\ 
    &\phantom{=\ \ \ }\mathbf{e}_{r^n b \cdot b}, \mathbf{e}_{r^n b \cdot rb}, \dots, \mathbf{e}_{r^n b \cdot r^{N-1} b}]  
\end{align}
using the group laws $b^2 = e$ and $r^n b = b r^{-n}$. By changing the order of cyclic rotation and reflection, the equation can be transformed as:
\begin{align}
\sigma^{\mathrm{D}_N}_{\mathrm{reg}}(b r^n)
    &= [\mathbf{e}_{b r^n}, \mathbf{e}_{b r^n \cdot r}, \dots, \mathbf{e}_{b r^n \cdot r^{N-1}}, \nonumber\\
    &\phantom{=\ \ \ }\mathbf{e}_{b r^n \cdot b}, \mathbf{e}_{b r^n \cdot rb}, \dots, \mathbf{e}_{b r^n \cdot r^{N-1} b}]  
\end{align}
The basis vectors for the dihedral group are defined from:
\begin{align}
\sigma^{\mathrm{D}_N}_{\mathrm{reg}}(r^0b^0) = \mathrm{I}_{2N}.
\end{align}

\subsection{Discrete group convolution}
\label{subsection: Discrete Group Convolution}
Conventional convolutional neural networks (CNNs) are inherently equivariant to translations, meaning that a translation of the input results in a corresponding translation of the output. The standard 2D convolution operation can be expressed as:
\begin{align}
\left(f * \psi\right)\left(\mathbf{x}\right) = \sum_{\mathbf{y} \in \mathbb{Z}^2} f\left(\mathbf{y}\right) \psi\left(\mathbf{x} - \mathbf{y}\right),
\end{align}
where $f: \mathbb{Z}^2 \rightarrow \mathbb{R}^{\mathcal{C}_\mathrm{in}}$ is the input function with $\mathcal{C}_\mathrm{in}$ channels, $\psi: \mathbb{Z}^2 \rightarrow \mathbb{R}^{\mathcal{C}_\mathrm{in} \times \mathcal{C}_\mathrm{out}}$ is the filter, and $\mathbf{x}, \mathbf{y} \in \mathbb{Z}^2$ are spatial coordinates. Here, plane feature map is defined only along the spatial dimension $\mathbb{Z}^2$. To associate discrete group within the feature map, an additional dimension corresponding to the group $G$ should be constructed, resulting in the mapping $f_G:G\times \mathbb{Z}^2 \rightarrow \mathbb{R}^\mathcal{C}$. In the discrete group convolution, this additional dimension is constructed through the lifting operation:
\begin{equation}
f_{G} = \bigoplus_{g\in G} \left(f\ast g\psi\right).
\end{equation}
The order of the stack corresponds to the order of group elements in the initial state. Since the lifted feature map contains features corresponding to each group element, transformations must account for both spatial changes and the group structure. Applying a specific group element $g' \in G$ to the lifted feature map thus requires both spatial transformation and permutation of the group dimension:
\begin{align}
(g' \cdot f_G)(\mathbf{x}) = \sigma^G_\mathrm{reg}(g') \cdot f_G( g'^{-1}\mathbf{x}),
\end{align}
where $\sigma^G_\mathrm{reg}(g')$ is the block diagonal form of the regular representation of $g'$ repeated $\mathcal{C}$ times, permuting along the group dimension, while $g'^{-1} \cdot \mathbf{x}$ applies the spatial transformation. Following the lifting operation, group convolution for the lifted feature map is defined as:
\begin{align}
&\left[f_G \ast_G \psi\right]\left(g, \mathbf{x}\right) \nonumber \\
&= \sum_{g' \in G} \sum_{\mathbf{y} \in \mathbb{Z}^2} f_G\left(g', \mathbf{y}\right) \left[\sigma^G_\textrm{reg}(g)\psi(g^{-1}(\mathbf{x}-\mathbf{y}))\right](g').
\end{align}

Here, $\psi: G\times\mathbb{Z}^2\rightarrow \mathbb{R}^{\mathcal{C}_\mathrm{in}\times \mathcal{C}_\mathrm{out}}$ represents the group convolution filter, where $f_G:G\times \mathbb{Z}^2\rightarrow \mathbb{R}^{\mathcal{C}_\mathrm{in}}$ is the lifted feature map, and $g, g' \in G$ are group elements of $G$. The key property of group convolution is its equivariance to group elements, expressed as:
\begin{align}
[(g' \cdot f_G) \ast_G \psi](\mathbf{x}) &= [g' \cdot (f_G \ast_G \psi)](\mathbf{x}) \nonumber \\
&= \sigma^G_\mathrm{reg}(g') \cdot (f_G \ast_G \psi)(g'^{-1} \cdot \mathbf{x}),
\end{align}
for any $g' \in G$. Here, $(g' \cdot f_G) \ast_G \psi$ represents the group convolution applied to the transformed input, while $g' \cdot (f_G \ast_G \psi)$ is the action of $g'$ on the result of the group convolution. This equality demonstrates that the order of applying group transformations and group convolutions is interchangeable, preserving the group structure throughout the network layers.

\section{Cyclic group-equivariance of the reflectional matching }
\addcontentsline{toc}{section}{Appendix B: \texorpdfstring{$\mathrm{C}_N$}{CN} Equivariance of the Reflectional Matching}
\label{appendix B}
\subsection{Cyclic group-equivariance of the single fiber reflectional matching}
Given a $\mathrm{D}_N$-equivariant feature map $\mathbf{F} \in \mathbb{R}^{\mathcal{C}|\mathrm{D}_N| \times H \times W}$ under the regular representation $\sigma_\mathrm{reg}$, we need to prove that $\mathbf{H}$ from Reflectional Matching without spatial expansion is equivariant to the cyclic group $\mathrm{C}_N$ with its element $r^k$:
\begin{align}
    &\bigoplus_{n=0}^{N-1} h\left(
        \sigma_\mathrm{reg}^{\mathrm{D}_N}(r^n)\mathbf{F}^{(0, k)}_\mathbf{x}, 
        \sigma_\mathrm{reg}^{\mathrm{D}_N}(b r^n)\mathbf{F}^{(0, k)}_\mathbf{x}
    \right)  \nonumber \\
    &= \sigma_{\mathrm{reg}}^{\mathrm{D}_N}(r^k) 
    \bigoplus_{n=0}^{N-1} 
    h\left(
        \sigma_\mathrm{reg}^{\mathrm{D}_N}(r^n)\mathbf{F}^{(0, 0)}_\mathbf{x}, 
        \sigma_\mathrm{reg}^{\mathrm{D}_N}(b r^n)\mathbf{F}^{(0, 0)}_\mathbf{x}
    \right), 
\end{align}
where $\mathbf{F}_\mathbf{x}^{(l,n)}$ is the fiber at position $\mathbf{x}$, with the regular representation corresponding to $l$ reflections and $n$ rotations added. Using the property $\sigma(g)\sigma(h) = \sigma(gh)$, the equation can be rewritten as:
\begin{align}
    &\bigoplus_{n=0}^{N-1} \, h\left(
        \sigma_\mathrm{reg}^{\mathrm{D}_N}(r^n)\mathbf{F}^{(0, k)}_\mathbf{x}, 
        \sigma_\mathrm{reg}^{\mathrm{D}_N}(b r^n)\mathbf{F}^{(0, k)}_\mathbf{x}
    \right)  \nonumber \\
    &= \bigoplus_{n=0}^{N-1} 
    h\left(
        \sigma_\mathrm{reg}^{\mathrm{D}_N}(r^{k+n})\mathbf{F}^{(0, 0)}_\mathbf{x}, 
        \sigma_\mathrm{reg}^{\mathrm{D}_N}(b r^{k+n})\mathbf{F}^{(0, 0)}_\mathbf{x}
    \right).
\end{align}
Here, $h$ is the similarity function defined as:
\begin{align}
    h(\mathbf{f}^1, \mathbf{f}^2) = \bigoplus_{c=1}^\mathcal{C} \frac{\mathbf{f}^1_c \cdot \mathbf{f}^2_c}{\| \mathbf{f}^1_c \| \| \mathbf{f}^2_c \|} \in \mathbb{R}^\mathcal{C},
\end{align}
Since permutation matrices preserve the norm of a vector, and using the rule $r^{a+b}=r^{(a+b)\bmod N}$, the equation can be reformulated as:
{\small
\begin{align}
    &\bigoplus_{n=0}^{N-1} \bigoplus_{c=1}^{\mathcal{C}} \, \frac{1}{\|\mathbf{F}_{c, \mathbf{x}}^{(0,0)}\|^2} \left( 
        \sigma_\mathrm{reg}^{\mathrm{D}_N}(r^{k+n})\mathbf{F}^{(0, 0)}_{c, \mathbf{x}} \cdot \sigma_\mathrm{reg}^{\mathrm{D}_N}(b r^{k+n})\mathbf{F}^{(0, 0)}_{c, \mathbf{x}} 
    \right) \nonumber  \\
    &= \bigoplus_{c=1}^{\mathcal{C}} \frac{1}{\|\mathbf{F}_{c, \mathbf{x}}^{(0,0)}\|^2} 
    \bigoplus_{n=0}^{N-1} \left( 
        \sigma_\mathrm{reg}^{\mathrm{D}_N}(r^{k+n})\mathbf{F}^{(0, 0)}_{c, \mathbf{x}} \cdot \sigma_\mathrm{reg}^{\mathrm{D}_N}(b r^{k+n})\mathbf{F}^{(0, 0)}_{c, \mathbf{x}} 
    \right)   \\
    &= \bigoplus_{c=1}^{\mathcal{C}} \frac{1}{\|\mathbf{F}_{c, \mathbf{x}}^{(0,0)}\|^2} 
    \bigoplus_{n=k}^{k+N-1} \left( 
        \sigma_\mathrm{reg}^{\mathrm{D}_N}(r^{n})\mathbf{F}^{(0, 0)}_{c, \mathbf{x}} \cdot \sigma_\mathrm{reg}^{\mathrm{D}_N}(b r^{n})\mathbf{F}^{(0, 0)}_{c, \mathbf{x}} 
    \right)   \\
    &= \bigoplus_{c=1}^{\mathcal{C}} \frac{1}{\|\mathbf{F}_{c, \mathbf{x}}^{(0,0)}\|^2} \sigma_\mathrm{reg}^{\mathrm{D}_N}(r^k) \nonumber \\
    &\phantom{= \sigma_\mathrm{reg}^{\mathrm{D}_N}(r^k)\ } \bigoplus_{n=0}^{N-1} \left( 
        \sigma_\mathrm{reg}^{\mathrm{D}_N}(r^{n})\mathbf{F}^{(0, 0)}_{c, \mathbf{x}} \cdot \sigma_\mathrm{reg}^{\mathrm{D}_N}(b r^{n})\mathbf{F}^{(0, 0)}_{c, \mathbf{x}} 
    \right)   \\
    &= \sigma_\mathrm{reg}^{\mathrm{D}_N}(r^k) 
    \bigoplus_{c=1}^{\mathcal{C}} \frac{1}{\|\mathbf{F}_{c, \mathbf{x}}^{(0,0)}\|^2} \nonumber \\ 
    &\phantom{= \sigma_\mathrm{reg}^{\mathrm{D}_N}(r^k)\ } \bigoplus_{n=0}^{N-1} \left( 
        \sigma_\mathrm{reg}^{\mathrm{D}_N}(r^{n})\mathbf{F}^{(0, 0)}_{c, \mathbf{x}} \cdot \sigma_\mathrm{reg}^{\mathrm{D}_N}(b r^{n})\mathbf{F}^{(0, 0)}_{c, \mathbf{x}} 
    \right)   \\
    &= \sigma_\mathrm{reg}^{\mathrm{D}_N}(r^k) 
    \bigoplus_{n=0}^{N-1} 
    h\left(
        \sigma_\mathrm{reg}^{\mathrm{D}_N}(r^n)\mathbf{F}^{(0, 0)}_\mathbf{x}, 
        \sigma_\mathrm{reg}^{\mathrm{D}_N}(b r^n)\mathbf{F}^{(0, 0)}_\mathbf{x}
    \right),
\end{align}
}
where $\mathbf{F}_{c,\mathbf{x}}$ denotes the feature at position $\mathbf{x}$ in channel $c$.

\subsection{Cyclic group equivariance of spatially expanded reflectional matching}
We now have to to prove the spatial expansion of single fiber Reflectional Matching is also equivariant to the cyclic group $\mathrm{C}_N$:
\small{
\begin{align}
    &\bigoplus_{n=0}^{N-1} \, \sum_{\mathbf{q}\in\mathcal{Q}} 
    h\left(
        \sigma_\mathrm{reg}^{\mathrm{D}_N}(r^n)\mathbf{F}^{(0, k)}_{\mathbf{x} + r^{k+n}(\mathbf{q})}, 
        \sigma_\mathrm{reg}^{\mathrm{D}_N}(b r^n)\mathbf{F}^{(0, k)}_{\mathbf{x} + b r^{k+n}(\mathbf{q})}
    \right)   \nonumber\\
    &= \sigma_{\mathrm{reg}}^{\mathrm{D}_N}(r^k) 
    \bigoplus_{n=0}^{N-1} 
    \sum_{\mathbf{q}\in\mathcal{Q}} 
    h\left(
        \sigma_\mathrm{reg}^{\mathrm{D}_N}(r^n)\mathbf{F}^{(0, 0)}_{\mathbf{x} + r^n(\mathbf{q})}, 
        \sigma_\mathrm{reg}^{\mathrm{D}_N}(b r^n)\mathbf{F}^{(0, 0)}_{\mathbf{x} + b r^n(\mathbf{q})}
    \right), 
\end{align}}
where $\mathbf{q} \in \mathcal{Q}$ is the offset, $r^n(\mathbf{q})$ represents the spatially rotated offset, and $b r^n(\mathbf{q})$ denotes the offset that is first rotated and then reflected. Same as single fiber, the equation can be written as:
{\small
\begin{align}
    &\bigoplus_{n=0}^{N-1} \, \sum_{\mathbf{q}\in\mathcal{Q}} 
    h\left(
        \sigma_\mathrm{reg}^{\mathrm{D}_N}(r^{k+n})\mathbf{F}^{(0, 0)}_{\mathbf{x} + r^{k+n}(\mathbf{q})}, \right. \nonumber \\
    &\phantom{\bigoplus_{n=0}^{N-1} \, \sum_{\mathbf{q}\in\mathcal{Q}} h(}
        \left.\sigma_\mathrm{reg}^{\mathrm{D}_N}(b r^{k+n})\mathbf{F}^{(0, 0)}_{\mathbf{x} + br^{k+n}(\mathbf{q})}
    \right) \nonumber \\
    &=\bigoplus_{n=0}^{N-1}\sum_{\mathbf{q}\in\mathcal{Q}} \bigoplus_{c=1}^{\mathcal{C}} \, \frac{
        \sigma_\mathrm{reg}^{\mathrm{D}_N}(r^{k+n})\mathbf{F}^{(0, 0)}_{c, {\mathbf{x} + r^{k+n} (\mathbf{q})}} \cdot} {\|\mathbf{F}_{c, {\mathbf{x} + r^{k+n} (\mathbf{q})}}^{(0,0)}\| \| \mathbf{F}_{c, {\mathbf{x} + br^{k+n} (\mathbf{q})}}^{(0,0)}\|} \nonumber \\
    &\phantom{=\bigoplus_{n=0}^{N-1}\sum_{\mathbf{q}\in\mathcal{Q}} \bigoplus_{c=1}^{\mathcal{C}} \, \frac{}{}} \cdot
        \sigma_\mathrm{reg}^{\mathrm{D}_N}(b r^{k+n})\mathbf{F}^{(0, 0)}_{c, {\mathbf{x} + br^{k+n} (\mathbf{q})}}  \\
    &=\bigoplus_{c=1}^{\mathcal{C}} \, \bigoplus_{n=k}^{k+N-1}\sum_{\mathbf{q}\in\mathcal{Q}} \frac{
        \sigma_\mathrm{reg}^{\mathrm{D}_N}(r^n)\mathbf{F}^{(0, 0)}_{c, {\mathbf{x} + r^n (\mathbf{q})}} \cdot} {\|\mathbf{F}_{c, {\mathbf{x} + r^n (\mathbf{q})}}^{(0,0)}\| \| \mathbf{F}_{c, {\mathbf{x} + br^n (\mathbf{q})}}^{(0,0)}\|} \nonumber \\
    &\phantom{=\bigoplus_{c=1}^{\mathcal{C}} \, \bigoplus_{n=k}^{k+N-1}\sum_{\mathbf{q}\in\mathcal{Q}} \frac{}{}} \cdot
        \sigma_\mathrm{reg}^{\mathrm{D}_N}(b r^n)\mathbf{F}^{(0, 0)}_{c, {\mathbf{x} + br^n (\mathbf{q})}}  \\
    &=\bigoplus_{c=1}^{\mathcal{C}} \,\sigma_\mathrm{reg}^{\mathrm{D}_N}(r^k)  \bigoplus_{n=0}^{N-1}\sum_{\mathbf{q}\in\mathcal{Q}} \frac{
        \sigma_\mathrm{reg}^{\mathrm{D}_N}(r^n)\mathbf{F}^{(0, 0)}_{c, {\mathbf{x} + r^n (\mathbf{q})}} \cdot} {\|\mathbf{F}_{c, {\mathbf{x} + r^n (\mathbf{q})}}^{(0,0)}\| \| \mathbf{F}_{c, {\mathbf{x} + br^n (\mathbf{q})}}^{(0,0)}\|} \nonumber \\
    &\phantom{=\bigoplus_{c=1}^{\mathcal{C}} \,\sigma_\mathrm{reg}^{\mathrm{D}_N}(r^k)  \bigoplus_{n=0}^{N-1}\sum_{\mathbf{q}\in\mathcal{Q}} \frac{}{}} \cdot
        \sigma_\mathrm{reg}^{\mathrm{D}_N}(b r^n)\mathbf{F}^{(0, 0)}_{c, {\mathbf{x} + br^n (\mathbf{q})}}  \\
    &=\sigma_\mathrm{reg}^{\mathrm{D}_N}(r^k) \, \bigoplus_{n=0}^{N-1} \sum_{\mathbf{q}\in\mathcal{Q}} \bigoplus_{c=1}^{\mathcal{C}} \frac{
        \sigma_\mathrm{reg}^{\mathrm{D}_N}(r^n)\mathbf{F}^{(0, 0)}_{c, {\mathbf{x} + r^n (\mathbf{q})}} \cdot} {\|\mathbf{F}_{c, {\mathbf{x} + r^n (\mathbf{q})}}^{(0,0)}\| \| \mathbf{F}_{c, {\mathbf{x} + br^n (\mathbf{q})}}^{(0,0)}\|} \nonumber \\
    &\phantom{=\sigma_\mathrm{reg}^{\mathrm{D}_N}(r^k) \, \bigoplus_{n=0}^{N-1} \sum_{\mathbf{q}\in\mathcal{Q}} \bigoplus_{c=1}^{\mathcal{C}} \frac{}{}} \cdot
        \sigma_\mathrm{reg}^{\mathrm{D}_N}(b r^n)\mathbf{F}^{(0, 0)}_{c, {\mathbf{x} + br^n (\mathbf{q})}}  \\
    &= \sigma_{\mathrm{reg}}^{\mathrm{D}_N}(r^k) 
    \bigoplus_{n=0}^{N-1} 
    \sum_{\mathbf{q}\in\mathcal{Q}} 
    h\left(
        \sigma_\mathrm{reg}^{\mathrm{D}_N}(r^n)\mathbf{F}^{(0, 0)}_{\mathbf{x} + r^n(\mathbf{q})}, \right. \nonumber \\
    &\phantom{= \sigma_{\mathrm{reg}}^{\mathrm{D}_N}(r^k) 
    \bigoplus_{n=0}^{N-1} 
    \sum_{\mathbf{q}\in\mathcal{Q}} 
    h(}
        \left.\sigma_\mathrm{reg}^{\mathrm{D}_N}(b r^n)\mathbf{F}^{(0, 0)}_{\mathbf{x} + b r^n(\mathbf{q})}
    \right).
\end{align}}

\end{document}